\documentclass[10pt,twocolumn,letterpaper]{article}

\usepackage{iccv}
\usepackage{times}
\usepackage{epsfig}
\usepackage{graphicx}
\usepackage{amsmath}
\usepackage{amssymb}
\usepackage{times}
\usepackage{epsfig}
\usepackage{graphicx}
\usepackage{amsmath}
\usepackage{amssymb}
\usepackage{algorithm}
\usepackage{algorithmic}
\usepackage{tabulary}
\usepackage{subfigure}
\usepackage{capt-of}
\usepackage{ctable}
\usepackage{multirow}
\long\def\ignorethis#1{}

\usepackage{color}
\definecolor{gray}{rgb}{0.35,0.35,0.35}
\definecolor{red}{rgb}{1,0,0}
\definecolor{dark-green}{rgb}{0,0.4,0}
\definecolor{dark-pink}{rgb}{0.78,0.08,0.52}
\definecolor{blue}{rgb}{0,0,1}
\definecolor{orange}{rgb}{1,0.55,0}
\definecolor{white}{rgb}{1,1,1}
\definecolor{black}{rgb}{1,1,1}
\definecolor{dark-brown}{rgb}{0.2,0.1,0}

\newcommand{\RNum}[1]{\uppercase\expandafter{\romannumeral #1\relax}}

\ifdefined\ShowNotes

\else
  \newcommand{\colornote}[3]{}
\fi

\newbox\jsavebox

\usepackage{diagbox}
\usepackage{amsmath}

\usepackage[pagebackref=true,breaklinks=true,letterpaper=true,colorlinks,bookmarks=false]{hyperref}

\iccvfinalcopy 

\ificcvfinal\fi

\begin{document}

\title{How Low Can We Go? Pixel Annotation for Semantic Segmentation}

\author
{
Daniel Kigli\\
Tel-Aviv University\\
{\tt\small danielkigli@mail.tau.ac.il}
\and
Ariel Shamir\\
Reichman University\\
{\tt\small arik@idc.ac.il}
\and
Shai Avidan\\
Tel-Aviv University\\
{\tt\small avidan@eng.tau.ac.il}
}

\maketitle
\ificcvfinal\thispagestyle{empty}\fi

\begin{abstract}
How many labeled pixels are needed to segment an image, without any prior knowledge?
We conduct an experiment to answer this question.

In our experiment, an Oracle is using Active Learning to train a network from scratch. The Oracle has access to the entire label map of the image, but the goal is to reveal as few pixel labels to the network as possible. We find that, on average, the Oracle needs to reveal (i.e., annotate) less than $0.1\%$ of the pixels in order to train a network. The network can then label all pixels in the image at an accuracy of more than $98\%$. 

Based on this single-image-annotation experiment, we designed an experiment to quickly annotate an entire data set. In the data set level experiment, the Oracle trains a new network for each image from scratch. The network can then be used to create pseudo-labels, which are the network predicted labels of the unlabeled pixels, for the entire image. Only then a data set level network is trained from scratch on all the pseudo-labeled images at once.

We repeat both image-level and data set level experiments on two very different, real-world data sets, and find that it is possible to reach the performance of a fully annotated data set using a fraction of the annotation cost.

\end{abstract}

\section{Introduction}

\begin{figure}[!ht]
\begin{center}
\begin{tabular}{ c }
\includegraphics[width=0.95\linewidth]{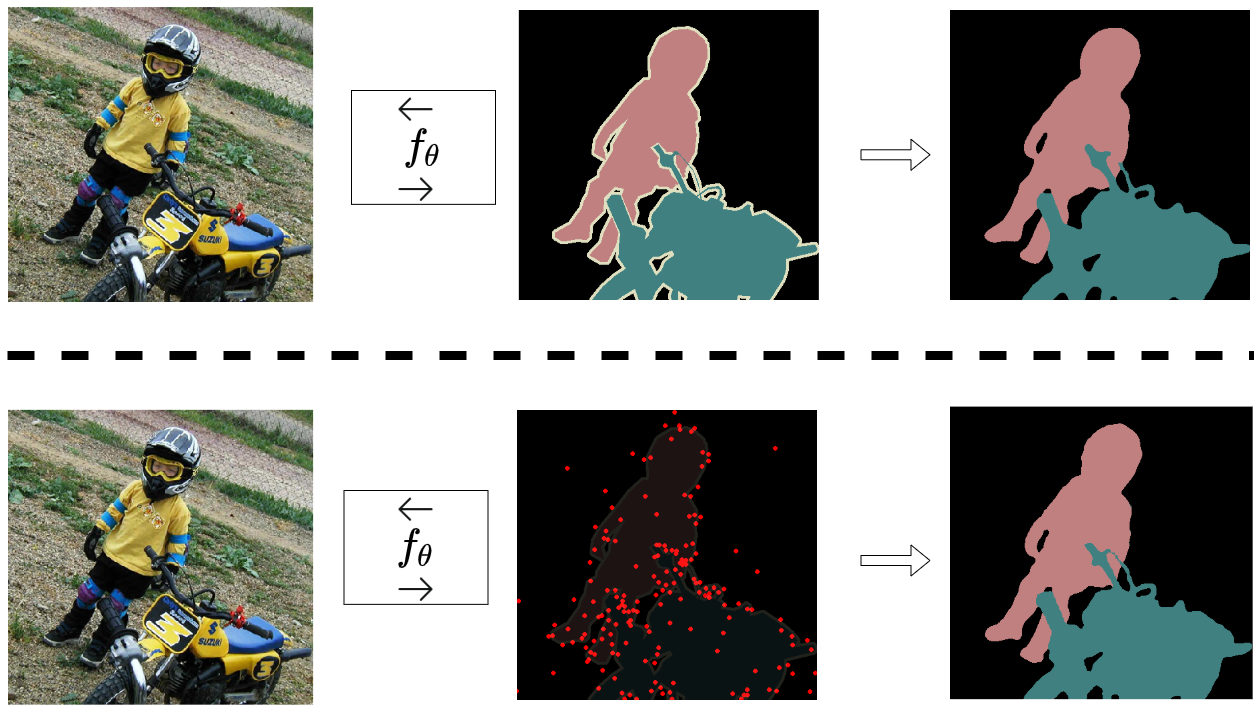}\\
\end{tabular}
\caption{{\bf Approximating the loss}. Training the network $f_{\Theta}(x)$ from scratch on a single image and its corresponding label map. On the right we see the label map produced by the trained network. (Top row) Full supervision: The loss is computed on all pixels in the label map. (Bottom row) The loss is computed on less than ~$0.1$\% of the labeled pixels (pixels marked in red). Both networks achieve comparable, and sometimes indistinguishable, results. On average, labeling $0.03\%$ pixels per image on Pascal VOC benchmark is enough to reach full-supervision results.}
\label{fig:teaser}
\end{center}
\end{figure}

Semantic segmentation of images is required in many computer vision applications such as image editing, image understanding, medical image analysis and more. Extensive research led to significant progress, yet this is still a challenge because of the large variability in image content.

Annotating a new data set of images is often based on interactively segmenting each image independently. For each image, the user interactively adds scribbles until she is satisfied with the segmentation proposed by the network. Repeating this process over many images creates a data set large enough to train a semantic segmentation network. 

As neural networks improve, the required number of user scribbles goes down. But these networks must first be trained on a data set with similar images for best performance, which leads to a chicken-and-egg problem. How do you bootstrap the process in case of a new data set of images that may not look similar to standard academic data sets?
This problem occurs in various sectors such as military, medical, or commercial.

So, how low can we go? what is the minimal number of labeled pixels that is required to train a semantic segmentation neural network from scratch? we conducted large scale experiments to answer this question. 

We first observe that when training a network on a single image, the loss can be computed on a subset of the pixels. See Figure~\ref{fig:teaser}. The network is trained from scratch on a single image with the loss defined only on the pixels marked with red. Once trained, the network can create a remarkably good pseudo-label for the entire image. This is in contrast to standard interactive image segmentation networks that are trained beforehand on a data set of similar images to perform semantic segmentation.

Equipped with this observation, we conducted our first experiment. In it, an Oracle, that has access to the entire label map of the image, uses Active Learning to train a network. At each step, the Oracle annotates some pixels (i.e., reveals their label), the network is trained on the available labeled pixels (i.e., the red pixels in Figure~\ref{fig:teaser}) and suggests a segmentation of the image. The Oracle compares the segmentation proposed by the network to the ground truth labels available to him, and provides additional annotations for training, until the network correctly labels the entire image.




We investigate three components that affect the training process. The first, of course, is the required number of labeled pixels and their location in the image plane. The second is whether we need to use pre-trained networks (i.e., a network that was pre-trained on ImageNet ~\cite{deng2009imagenet}), as opposed to initializing network weights with random values. Lastly, we also examine the importance of the order in which labeled pixels are introduced during training.





Next, we conduct a data set level experiment to estimate the amount of annotation required to train a network on an entire data set. A na\"ive extension of our findings above would be to annotate $0.1\%$ of the pixels in many, or all of the images, and use only them to train a network for the entire data set. However, we found that this did not work well, and took a different approach. 

Instead, we use the single-image-annotation approach to annotate images. That is, the Oracle trains a new network from scratch for each image, and the trained network is used to create pseudo-labels for the entire image. Once enough images are annotated this way, we train a new network for the entire data set. 


We found that, on average, the Oracle annotates less than $0.05\%$ of the pixels per image to reach full-supervision results. These conclusions also hold for a low-data regime where we use just a subset of the training images.

We conducted these experiments on two very different types of data sets: the Pascal-VOC 2012 data set ~\cite{pascal-voc-2012}, a popular instance segmentation benchmark, and Kvasir-SEG ~\cite{jha2021comprehensive}, a medical imaging binary segmentation set. 
We also ran an experiment on a large subset of images from the long-tailed distribution ADE20K data set~\cite{zhou2017scene}, and observed similar results, although a bit higher (~$0.5\%$), as it contains significantly higher class diversity in each image, so more annotations are needed. See supplemental for more details about this data set.

\paragraph{Why bother?} Why bother conducting these experiments? There are a couple of possible reasons. First, they give an unbiased estimate of how low we can go. Using networks that were trained on external data sets introduces a bias that is difficult to measure (i.e., what is the correlation between the external data set and the new data set to be annotated?). Second, they will let us estimate if current user-based systems can go that low. We might also learn how to instruct users to annotate images faster. Finally, they offer a test bed for current Active Learning algorithms to go even lower.

\newcommand{\mysizey}{0.96}
\newcommand{\mysizeyKvasir}{0.2}

\begin{figure*}[htb!]

\centering
\begin{tabular}{c}

\includegraphics[width=\mysizey\linewidth]{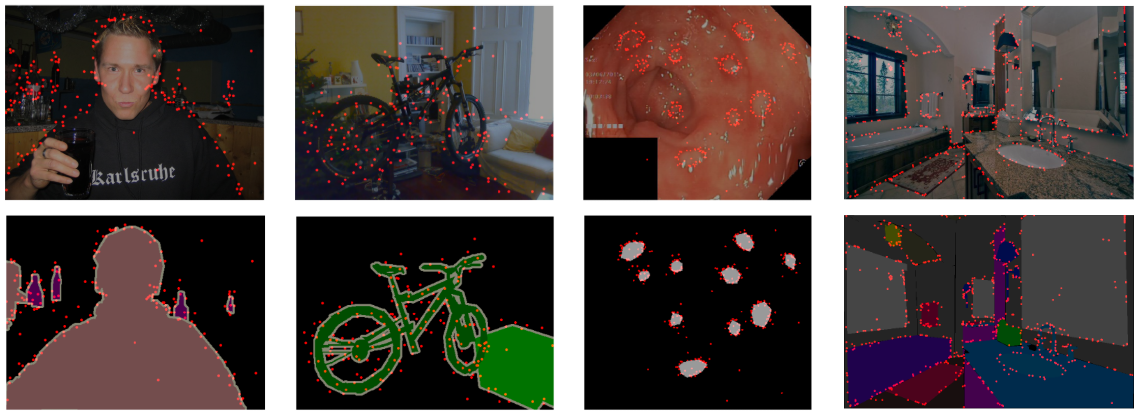} \\


\end{tabular}
\caption{Visualization of the sampled points, achieved using Algorithm~\ref{algo:pseudo_sample}, over the image and labels. We can see that most sampled points are around the edges defined by the label image.}
\label{fig:examples_sampled_points_pseudo_optimal_narrow}
\end{figure*}

\section{Related Work}
The literature on semantic segmentation is vast and only work directly related to ours will be covered here.
Supervision on the task of semantic segmentation usually involves using Cross-Entropy (CE) term for each labeled pixel on the CNN output map, which involves creating expensive annotations for the entire image.

There has been considerable work on reducing the amount of labor involved in semantically labeling an entire data set.

\paragraph{Weakly-Supervised Segmentation}
Instead of labeling each pixel in the training set, one can use weakly supervised methods. These includes image-level labels~\cite{vernaza2017learning, huang2018weakly}, clicks~\cite{BearmanRFL16}, and scribbles~\cite{Lin2016ScribbleSupSC, tang2018normalized, tang2018regularized, wang2019boundary}, where \cite{Lin2016ScribbleSupSC} provided scribbles data set on Pascal-VOC that include about $3\%$ of labeled pixels. However, all those scribble-based approaches did not close the performance gap compared to full-supervision. We, on the other hand, use two orders of magnitude less supervision ($0.03\%$) while reaching full-supervision results.

\paragraph{Interactive Segmentation}
Annotating a data set usually relies on interactive image segmentation tools, often geared towards foreground/background separation. Common approaches \cite{Xu_2016_CVPR,li2018interactive,agustsson2019interactive,lin2020interactive} differ in the number of user-clicks required, generalization to unseen datasets and the amount of supervision needed for training. On the other hand, we are interested in multi-label segmentation without any prior knowledge. Moreover, while interactive segmentation relies on refining trained models at inference time, we train our network \emph{from scratch} at inference time.



\paragraph{Active Learning Sampling}
Our work is related to Active Learning that iteratively selects the most informative samples to label~\cite{Settles}. In one popular variant, uncertainty pool-based active learning, softmax probabilities from the model output are used to compute entropy as an uncertainty criterion for future manual annotation~\cite{Gal2017}. Recently, Siddiqui {\em et al.}~\cite{SiddiquiVN20} proposed ViewAL that learns semantic segmentation in multi-view datasets by exploiting viewpoint consistency. This allowed them to use just $7\%$ of the pixels to label all pixels in the dataset. Casanova {\em et al.} \cite{casanova2020reinforced} showed that using labels of $9\%$ of CityScapes~\cite{Cityscapes} dataset can achieve full-supervision results by labeling only small informative image regions.

We make use of Curriculum Learning in our work. Curriculum Learning was shown by Bengio {\em et al.} \cite{bengio2009curriculum} to increase network performance and generalization, by presenting the network with examples that gradually increase in complexity, and not in random order. They showed this at the \emph{image-level}, while we showed similar behavior at the \emph{pixel-level}. Training in a pre-defined order can improve results compared to training without any order, and training in an inverse order produce the lowest results.

Can some pixels be considered ''easier'' or ''harder''? Li {\em et al.} \cite{li2017not} introduced such a notion and showed a small improvement in the results. However, their measurement is based on the ambiguity of the network. They classified a large portion of the image as ''moderate'' or ''hard'' difficulty as well as using additional datasets.
We, on the other hand, can learn without prior knowledge, for each image separately, and show considerable gains.

Some work explores redundancies of images in a dataset. Birodkar {\em et al.} \cite{birodkar2019semantic} explored such redundancies in image classification and found that on common datasets, one can ignore $10\%$ of the training data, and still reach the same performance. We show that in dense semantic segmentation scenarios, we can ignore $99.9\%$ of the labeled pixels while reaching the same results.

\vspace{-0.2cm}
\paragraph{Zero-Shot Semantic Segmentation}
Segmenting images where not all classes are represented in the training examples, can be done by transferring knowledge to unseen classes, as shown by Bucher {\em et al.} \cite{bucher2019zero} that generated synthetic training data for unseen classes.
However, we are interested in learning without any prior knowledge, at inference time. It was recently shown by Gandelsman {\em et al.}~\cite{gandelsman2019double} that it is possible to segment an image, without any prior knowledge into foreground and background. However, this does not deal with multi-label segmentation, nor does it encode user inputs to refine prediction.

\begin{figure*}[ht!]
\begin{center}
\begin{tabular}{ c }
\includegraphics[width=0.97\linewidth]{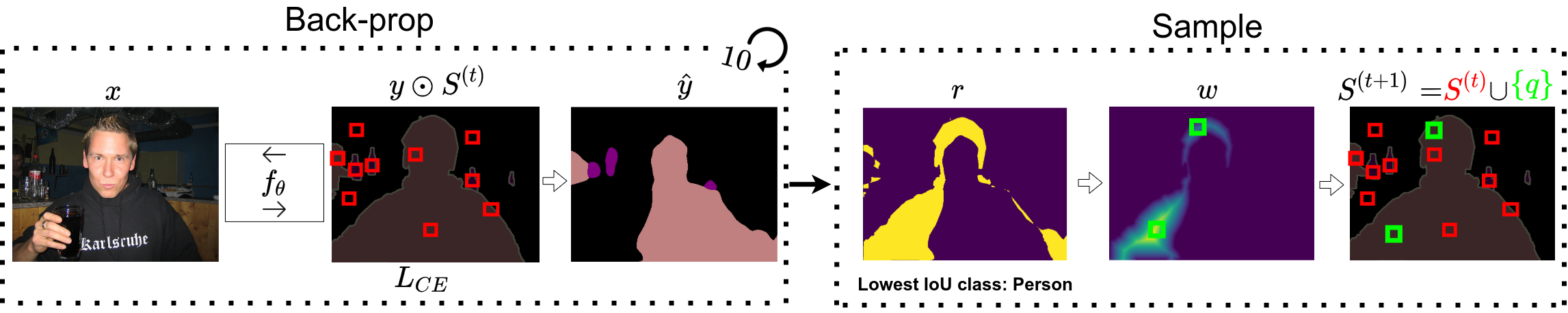} \\
\end{tabular}
\caption{Visualization of a single iteration from Algorithm~\ref{algo:pseudo_sample}. In each iteration we use the Oracle to select the next most informative point(s) as detailed in the algorithm. At the first stage, we fine-tune the network with the current queried points, $S^{(t)}$, for 10 iterations. At the second stage, we find the next point(s) to sample.}
\label{fig:method::algo}
\end{center}
\end{figure*}

\section{Single Image Annotation}
We conduct an experiment for the semantic annotation of an entire data set. The experiment consists of two parts: a single image annotation algorithm, and a data set level annotation step. In this section we focus on the former, and in the next section we will focus on the latter.

\subsection{Active Learning} 

We use a standard active learning approach to annotate an image. In our setting, the user is replaced with an Oracle that has access to the ground truth label map. At each step, the Oracle samples a few more pixels and reveals their labels to the CNN. The CNN is then trained on the entire image, but the loss is defined only on the pixels revealed so far by the Oracle. After a few back-propagation iterations, the Oracle samples a few more pixels, reveals their label to the CNN, and the process iterates. See outline in Figure~\ref{fig:method::algo}.

Given image $x$ with associated label map $y$ that consists of up to $C$ different class labels, we wish to learn a network $f_\Theta$ with parameters $\Theta$ such that:
\begin{equation}
    {\hat y} = f_\Theta(x)
\end{equation}
with loss function:
\begin{equation}
    L = \sum_{s=1}^{|S|} CE({\hat y}_{p_s},y_{p_s})
\end{equation}
where $CE$ is the Cross-Entropy loss, and $S$ is the set of indices of labeled pixels.
We are interested in scenarios in which the number of labeled pixels $|S|$ is much smaller than the number of pixels in the image $|x|$ (i.e., $|S| \ll |x|$).


The Oracle should sample points so as to reduce our goal, which is set to be the mean Intersection over Union (mIoU):
\begin{equation}
    mIoU = \frac{1}{C} \sum_{c=1}^{C} IoU_c
\end{equation}
where the IoU for a class is:
\begin{equation}
    IoU = 100 \frac{TP}{TP+FN+FP}
\end{equation}
and $TP, FP, FN$ are True Positive, False Positive, and False Negative, respectively.

The sampling is based on several criteria: a pixel should be with a label of the lowest IOU score class, it should be of the largest misclassified connected component, and it should be as far away as possible from the correctly classified pixels. 

Given the current prediction ${\hat y}$ of the network, the Oracle samples from a weight map that is constructed as follows. Let $r$ denote the misclassification map:
\begin{equation}
    r_q = \mathbf{1}[{\hat y}_q \neq y_q]
\end{equation}
and denote $R$ the set of pixels that belong to the largest connected component blob in $r$. Define the weight image $w$:
\begin{equation}\label{eq:compute_weight_map}
    w_q = \left\{ \begin{array}{ll}
                        min\{d(q, p) | p \notin R \} & q \in R \\
                        0 & \mbox{Otherwise}
                 \end{array} \right.
\end{equation} 
Where $d(p,q)$ is defined as the Euclidean distance. We then normalize $w$ such that $\sum_p{w_p}=1$. In other words, we prefer to sample points that are near the center of the largest misclassified blob. After a few iterations, the misclassified regions are mostly near the object boundaries. The Oracle then samples the next few pixels from $w$ according to their weight.



We have used the mIoU as our measure of success, but changing our goal would require changing just a few lines in the algorithm. For example, we can replace the way we compute the weight map $w$ with a different calculation that maximizes the pixel accuracy by sampling the center pixel of the largest misclassified region.

 \begin{algorithm}[h]
 \begin{algorithmic}[1]
 \renewcommand{\algorithmicrequire}{\textbf{Input:}}
 \renewcommand{\algorithmicensure}{\textbf{Output:}}
 \REQUIRE Image $x$ ; Segmentation mask $y$ ; Scalar $k$ ; Threshold T
 \ENSURE  Selected set of pixels to use for training
 \newline
  \STATE pick a random pixel $p \in y$
  \STATE Initialize: Segmentation network $f_{\theta}$ ; \newline $S^{(1)} = \{p\} $ ;  $t = 1$ \label{op::initialize_network}
  \REPEAT
  \FOR{$i \in \{1,\dots,10\}$} 
        \STATE $\mbox{minimize}_{\Theta}$ $L_{CE}(f_{\theta}(x),y\odot S^{(t)})$ \label{op::training}
  \ENDFOR
  \STATE Compute normalized weight map $w$ according to Equation~\ref{eq:compute_weight_map} \label{op::create_weight_map}
  \STATE Sample the next pixel(s) $q$ by weighted random sampling using $w$ \label{op::weighted_random_sampling}
  \STATE $S^{(t+1)} = S^{(t)} \cup \{q\}$ ; $t=t+1$
  \UNTIL{$|S^{(t)}| \geq k$ \OR $mIoU(f_{\theta}(x), y) \geq T$} \label{op::stoppping_criteria}
   \RETURN $S^{(t)}$ 
 \end{algorithmic}
 \caption{PixelLabeling($x, y, k, T$). Image-Level sampling algorithm. In the first phase, we train the network based on the current sampled labels for $10$ iterations, while in the second phase, we aim to find the most \emph{informative} point(s) to sample. We return the queried pixels as well as their order.}
 \label{algo:pseudo_sample}
 \end{algorithm}

\paragraph{Data sets} We evaluate the experiment on two data sets from different domains: Pascal-VOC 2012~\cite{pascal-voc-2012} and Kvasir-SEG~\cite{jha2021comprehensive}. Pascal-VOC semantic segmentation contains 20 classes and has 1464 images for training and 1449 for validation, and the validation set is used for testing. Kvasir-SEG is a medical binary segmentation data set for polyp segmentation. It contains 1000 images, and we randomly selected 800 images for training and 200 images for testing.

\begin{figure*}[h!]
\begin{center}
\begin{tabular}{c c}
\includegraphics[width=0.47\linewidth]{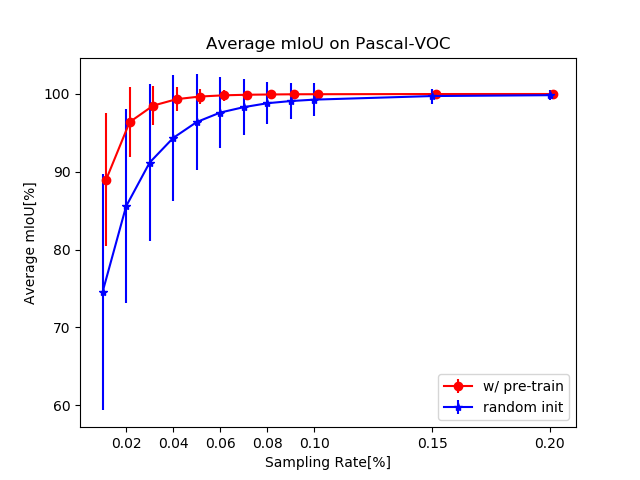}& \includegraphics[width=0.47\linewidth]{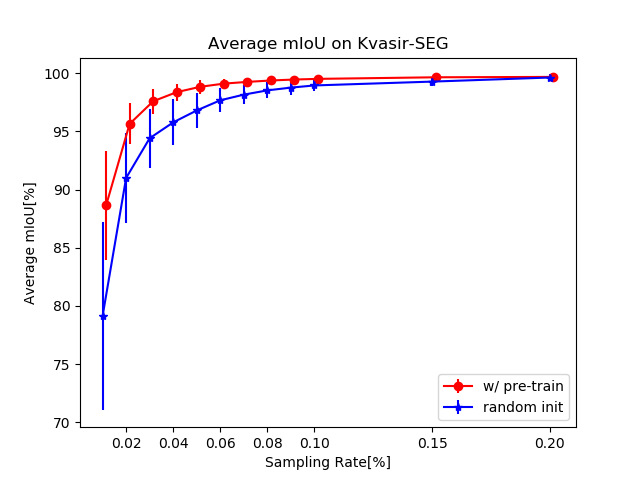}\\
\end{tabular}
\caption{Image-Level visualization - Resulting mIoU scores and variance (bars), for different sampling rates, obtained using Algorithm~\ref{algo:pseudo_sample}, and averaged across all training images. We compare a randomly initialized network results to an ImageNet pre-trained one, and show results for both Pascal-VOC and Kvasir-SEG data sets.}
\label{fig:with_and_without_imagenet}
\end{center}
\end{figure*}

\paragraph{Implementation Details}
We validate our results on several different segmentation architectures, but we show our results on DeepLabV3+~\cite{ChenZPSA18} as it is more widely used and does a better job at encoding spatial information. We set the output stride to 16 and use Resnet101 as the backbone, that can be either pre-trained on ImageNet, or randomly initialized, as the rest of the parameters in the model. We use standard Cross-Entropy (CE) loss on the labeled pixels, and train with an Adam optimizer with a learning rate of $2e-4$ and weight decay of $1e-4$.

As we have a highly sparse label image, we wish to extract most of what we are given by the Oracle. For this reason, step~\ref{op::training} in Algorithm~\ref{algo:pseudo_sample} consists of several iterations of forward-backward.
We observed that sampling up to $10$ pixels, instead of just one pixel, in step~\ref{op::weighted_random_sampling} speeds up the algorithm with little degradation in overall performance. Also, we set the threshold $T$ to $99.5\%$ to reduce potential clicks that do not improve the final result by much.

Our runtime depends on the network architecture, image size, the number of forward-backward iterations, and the number of pixels we sample in each iteration. On average, it takes $1$ minute per image on a standard GPU (GTX 1080 Ti).

Figure~\ref{fig:examples_sampled_points_pseudo_optimal_narrow} shows some examples of the points sampled by our algorithm. Visually examining the locations of the sampled points across the images shows that most of the points are near the boundaries of objects (according to the ground truth label image). We conclude that the common intuition of sampling in the proximity of object boundaries does hold. 

Naively using a classic method of sampling around the object boundaries did not produce good results, even when fused with grid sampling to cover more areas (see supplemental). We can attribute it to the fact that it is not clear in advance which edge pixels are important and where exactly we should sample the edge pixels, closer or farther from the edge. We argue that all of this is highly dependent on the state of the network. For extensive comparison to classic sampling methods, please see the supplemental. 




\subsection{Main Results} 

We conduct three experiments to evaluate our results. The first measures the number of labeled pixels that are required to properly train the network for a single image. The second is designed to measure the importance of using pre-trained features. The third experiment measures how the order in which labeled pixels are revealed during training affects network accuracy.
Additional results from ADE20K data set can be found in the supplemental.

\newcommand{\mysize}{0.18}

\begin{figure*}[h!]
\begin{center}
\begin{tabular}{c c c c c}

\textbf{Correct Order} & \textbf{No Order}  & \textbf{Random Order} & \textbf{Reverse Order}  \\
\includegraphics[width=\mysize\linewidth]{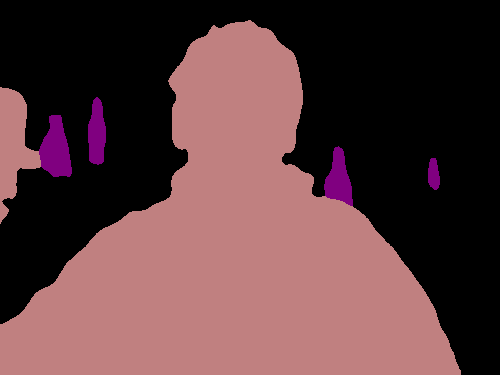} &
\includegraphics[width=\mysize\linewidth]{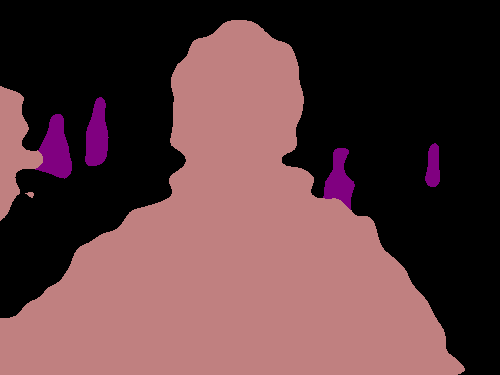} &
\includegraphics[width=\mysize\linewidth]{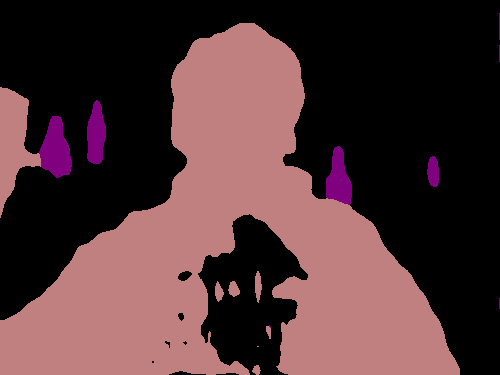} &
\includegraphics[width=\mysize\linewidth]{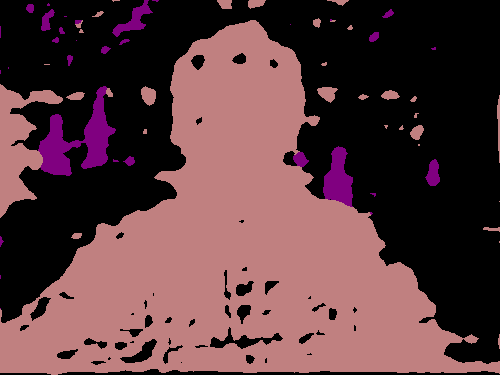} \\

\includegraphics[width=\mysize\linewidth]{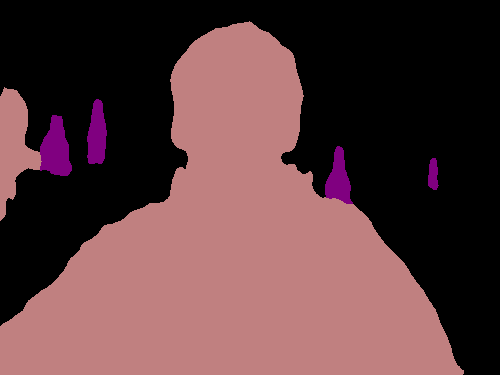} &
\includegraphics[width=\mysize\linewidth]{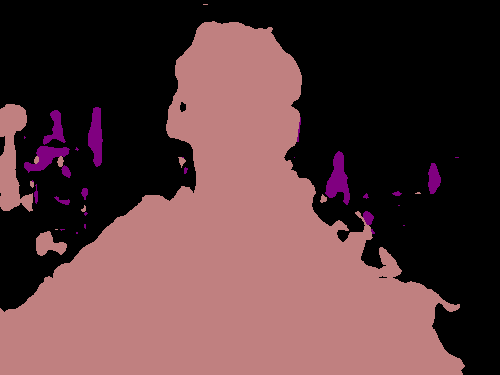} &
\includegraphics[width=\mysize\linewidth]{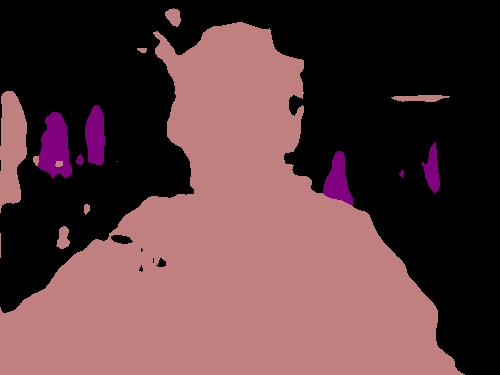} &
\includegraphics[width=\mysize\linewidth]{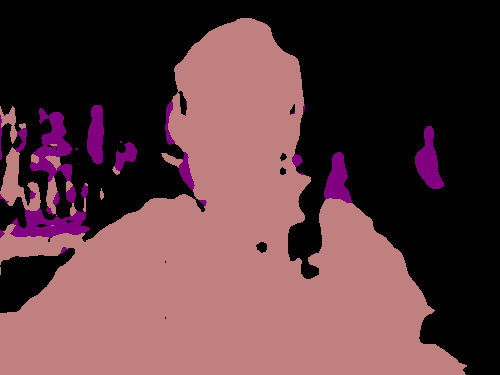} \\

\end{tabular}
\caption{{\bf Order and Initialization:} Results on a single image using different orders for training and different initialization. (Top) results using pre-trained initialized networks. (Bottom) results using randomly initialized networks. Each image represents a different network prediction, using one of four different network training procedures over a sparse 0.1\% labeled image. Both types of networks (pre-trained/random) produce visually pleasing results in the case of "Correct Order". In the case of "No Order", the pre-trained initialized network works better. Both types fail in the Reverse and Random Order regimes.}
\label{fig:results::different_training_with_without_imagenet}
\end{center}
\end{figure*}

\paragraph{Required number of points} \label{section:required_num_of_points}
We run our single-image-annotation algorithm (Algorithm~\ref{algo:pseudo_sample}) on all images in the training set of both data sets, where a new network is trained for each image independently. We compute the value $k$ as a percentage of the number of pixels in the image and compare various sampling rates other than $0.1\%$. 

Figure~\ref{fig:with_and_without_imagenet} (the red plots) shows the result. In both data sets, we achieve mIoU score of around $99.5\%$ with as little as $0.1\%$ of the pixels. That is, our experiments show that using about $200$ pixels, or roughly $0.1\%$ out of a $200,000$ pixel image, is enough to produce results that approach those achieved when training with the labels of all the pixels in an image.  

In practice, we observe that, on average, the number of labeled pixels is much lower than $0.1\%$, due to the effect of the threshold score $T$. We need about $0.06\%$ labeled pixels to reach $99.99\%$ accuracy, and with less than $0.03\%$ we reach $99\%$. In Section~\ref{section:dataset_level}, we will show that using this rate, we can reach the same results as the full-supervision, when training a single network on the entire data set.


To put things in context, recent work on interactive image segmentation achieves a lower pixel count. Li {\em et al.} \cite{li2018interactive} reached $85\%$ and $90\%$ mIoU scores on Pascal-VOC using 7 and 10 clicks, respectively, by learning binary predictions for each class, and refining the results from the user clicks. We, on the other hand, require about twice the number of clicks (roughly $15$ and $20$ clicks, respectively). Recently, \cite{lin2020interactive} showed further impressive results using less than 4 user clicks. However, they also trained their model using a data set that is similar to the one used for testing.

We do not compete with these methods, as we aim to answer a different question. Our goal is to determine how many clicks are required to train a network from scratch. Their work relies on models trained on large data sets, while in inference time, they only refine the output using the user clicks. We, on the other hand, train our network \emph{for scratch} for each image, at inference time, without any progressive learning. 
Another important distinction between us and prior work is that they focus on foreground/background segmentation, whereas we consider multi-label segmentation tasks. This automatically increases the number of required point clicks.

\paragraph{Rethinking ImageNet Pre-training}
Inspired by the work of \cite{he2019rethinking, zoph2020rethinking}, we aim to evaluate the contribution of an ImageNet pre-trained backbone on our results. We repeated the previous experiment, but this time using a backbone with randomly initialized weights (meaning that all network parameters were randomly initialized for each image). As can be seen (Figure~\ref{fig:with_and_without_imagenet}, blue plot), the mIoU drops only a little. For example, it drops by almost $10\%$ for the extremely low sampling rate of $0.01\%$ in the Pascal-VOC data set, from about $84\%$ (with pre-trained features) to about $76\%$ (without). The gap narrows to about $2\%$ at $0.05\%$ sampling rate. Similar behavior is observed in the Kvasir-Seg data set (right panel of Figure~\ref{fig:with_and_without_imagenet}).

As can be seen, using pre-trained features reduces the variance, but the gap quickly narrows as the sampling rate achieves $0.1\%$ and almost completely diminishes for a higher sampling rate. While \cite{he2019rethinking, zoph2020rethinking} showed that the contribution of pre-trained features diminishes for larger amounts of data, we observe a similar behavior at the pixel-level. Labeling enough pixels reduces the need for pre-training. On the other hand, compared to the number of resources needed to train large models from scratch (e.g., large batch sizes, heavy augmentation), using a \emph{single-image} model is much easier to train and does not involve large amounts of resources. 

\begin{table*}[h]
\centering
\begin{tabular}{l | c c |c c }
\hline
\multirow{3}{*}{\textbf{Method}} & \multicolumn{2}{c}{\textbf{Randomly Initialized}} & \multicolumn{2}{c}{\textbf{Pre-Trained}}\\
          & {\underline{Pascal-VOC}} & \underline{Kvasir-SEG} & {\underline{Pascal-VOC}} & \underline{Kvasir-SEG}\\
        \hline
No Order & {$76.02 \pm 17.61$} & {$86.68 \pm 7.70$} & {$94.79 \pm 5.92$} & {$98.46 \pm 2.64$} \\
\hline
Random Order & {$90.15 \pm 8.55$} & {$87.78 \pm 7.86$} & {$94.69 \pm 7.09$} & {$92.61 \pm 6.36$} \\
\hline
Inverse Order & {$80.88 \pm 11.42$} & {$71.77 \pm 10.85$} & {$81.06 \pm 12.52$} & {$64.42 \pm 10.79$}\\
\hline
\textbf{Correct Order} & $\mathbf{99.13 \pm 1.90}$ & $\mathbf{98.97 \pm 0.49}$ & $\mathbf{99.76 \pm 0.79}$ & $\mathbf{99.52 \pm 0.27}$ \\
\hline
\multicolumn{4}{c}{} 
\end{tabular}\\
\caption{Distribution of mIoU results over all training set, for the two data sets. Results (mean $\pm$ variance) are shown for each different network training procedure using a pre-trained/randomly initialized network.}
\label{tab:method::order_of_training}
\end{table*}

\paragraph{Effect of Samples Ordering}
In this section, we test whether the order in which labeled pixels are presented to the training process matters. To this end, we evaluated the following four different orderings:

\begin{enumerate}
\item{\bf No Order} Train the network on all of the pixels without any consideration to the order. In this configuration, we compute the loss on the entire labeled pixels and repeat the forward-backward pass for 200 iterations.
\item{\bf Correct order} Train the network in the same order as the pixels were added. Meaning that, in each iteration, we add the current pixel(s) to the sampling mask and compute the loss only on the pixels defined in the mask.
\item {\bf Random order} Train the network in a random order which is different from the order the pixels were added.
\item {\bf Reverse order} Train the network in the reverse order of the order they were added. This could be thought of as a variant of anti-curriculum ordering, instead of starting from the ``easy'' pixels, we start from the ``hard'' ones.
\end{enumerate} 

We repeated this experiment four times on all the training images in the data set, for pre-trained and randomly initialized networks,  and each order. For each image, we measured the resulting mIoU score achieved by each order.
In all cases, we used $0.1\%$ of the pixels as labels.

The results shown in Table~\ref{tab:method::order_of_training} indicate that order matters. Training in the correct order yields the best score with the lowest variance. Training with no consideration to ordering demonstrates a noticeable drop in accuracy and a large increase in the variance. Feeding the labeled pixels in random order hurts performance as well, and feeding the pixels in the reverse order leads to the lowest score and the largest variance. Changing hyper-parameters, like learning rate and amount of iterations did not produce better results. Thus, we conclude that the order pixels are presented matters.

Curriculum Learning was shown by \cite{bengio2009curriculum} to increase networks performance and generalization, by presenting the network with gradually increasing complex examples and not in random order. While previous works (e.g.\ \cite{vodrahalli2018all} ) show this at the \emph{image-level}, we show a similar behavior at the \emph{pixel-level}. Surprisingly, training in a pre-defined order can improve results compared to training without any order (i.e., on all available labels), and training in an inverse order produce the lowest results.

Similar to us, \cite{li2017not} introduced a notion of ''easy'' and ''hard'' pixels for the task of semantic segmentation. However, their measurement is based on the ambiguity of the network. They classify a large portion of the image as ''moderate'' or ''hard'' difficulty as well as using additional data sets. We, on the other hand, learn without any prior knowledge, for each image separately, and find a highly sparse (over two orders of magnitude less) subset of pixels.

The results of this experiment indicate a limitation of the training procedure, and not the data itself. This is an effect of the non-convexity of the problem and the stochastic nature of training neural networks, as there is no reason to believe that using the full set of labeled pixels will produce results that are inferior to training with the correct order.

Figure~\ref{fig:with_and_without_imagenet} shows the impact of the two types of networks (ImageNet pre-trained and randomly initialized), while Figure~\ref{fig:results::different_training_with_without_imagenet} visually shows the resulting label image achieved with different training procedures. As can be seen, the ImageNet pre-trained network tends to produce more visually pleasing results, while the randomly initialized network does so only in the case of the ''Correct Order''. Although the graphs in Figure~\ref{fig:with_and_without_imagenet} show a seemingly small gap, there is quite a large visual difference (the label map is noisier with distorted boundaries). This phenomenon was seen across all images, and more examples can be seen in the supplementary.

\begin{table*}[h]
\centering
\begin{tabular}{c | c c c c |c c c c}
\hline
\multirow{3}{*}{\textbf{\% Images}} & \multicolumn{8}{c}{\textbf{\% Labeled Pixels}}\\
          & \multicolumn{4}{c}{\underline{Pascal-VOC}} & \multicolumn{4}{c}{\underline{Kvasir-SEG}}\\
         & $0.01$\% & $0.05$\% & $0.1$\% & $100$\% & $0.01$\% & $0.05$\% & $0.1$\% & $100$\%\\
        \hline
10\% & 60.1 & \textbf{63.7} & 62.9 & 62.9 & 79.9 & 80.4 & 79.2 & \textbf{80.7}\\
\hline
30\% & 65.5 & \textbf{70.8} & 70.4 & 70.1 & 84.4 & \textbf{85.1} & 84.3 & 84.8\\
\hline
50\% & 65.9 & 72.9 & \textbf{73.1} & 72.1 & 86.9 & 87.1 & \textbf{87.2} & 86.9\\
\hline
100\% & 69.5 & 75.2 & \textbf{75.4} & 75.1 & 87.9 & 88.6 & 88.5 & \textbf{88.7}\\
\hline
\multicolumn{8}{c}{} 

\end{tabular}\\

\caption{{\bf Data set level performance:} mIoU results on the same test set, when using various fractions of labeled pixels per image and number of images for training. For example, in the Pascal-VOC data set, annotating $100\%$ of the pixels in $100\%$ of the training images gives $75.1\%$ mIoU on the test set. Using $0.05\%$ on $30\%$ of the images gives $70.8\%$ mIoU on the same test set. These results are based on the ImageNet pre-trained network.
}
\label{tab:dataset_level_training}
\end{table*}


\section{Data Set Level Training} \label{section:dataset_level}

Equipped with the results of the single image experiment, we turn to train a network on an entire data set. A na\"ive extension of our approach is to let the Oracle annotate a few pixels in every image and then train a network at once on all images with these annotations. However, we found that this does not work well. Training with just $0.1\%$ annotated pixels of $100\%$ of the training set images, leads to a drop of about $25\%$ in accuracy, compared to the one achieved using full supervision.

Instead, following previous work \cite{lee2013pseudo, bellver2019budget, xie2020self, zoph2020rethinking}, we use a simple self-training method that consists of three stages. First, the Oracle uses the single-image-annotation algorithm on a subset of the training images. Then, we train a model on the resulting images and use the trained model to create pseudo-labels on the remaining, unlabeled, images in the training set. Finally, a new model is trained over all labeled and pseudo labeled images in the training set.

Based on Figure~\ref{fig:with_and_without_imagenet} we found the $0.05\%$ sampling rate threshold for the single-image-network to give a good trade-off between sampling rate and segmentation accuracy. In practice, as explained in Section~\ref{section:required_num_of_points}, we observe that, on average, the number of labeled pixels is lower than $0.05\%$ and stands at about $0.03\%$. 

To avoid bias in the mean IoU calculations, we select images from the training data set for our experiment in such a way that preserves the distribution of classes found in the full training data set.

We report the results in Table~\ref{tab:dataset_level_training} as a function of two variables, the percentage of training set size (i.e., number of training images that are being used) and the supervision percentage per image (i.e., number of labeled pixels per image). 
We find that training with pseudo-labels works much better than simple na\"ive sparse label map training, and shows comparable performance to training with the true segmentation labels.

The results in Table~\ref{tab:dataset_level_training} were obtained using an ImageNet pre-trained network. We tried repeating the experiment with a randomly initialized network, and saw that the 0.1\% is sufficient to reach the same performance as the full-supervision. However, we saw a drop of about $45\%$ in performance. This phenomenon was also observed by ~\cite{zoph2020rethinking} that reached $28\%$ using a randomly initialized model on Pascal-VOC, as well as~\cite{he2019rethinking} who noticed that learning from scratch in a low-data regime leads to a large drop in performance. We can also note that training a model on an entire data set, without any pre-training, requires large computational efforts (e.g., huge batch sizes, extensive augmentations, efficient network).


We can further define the minimal sampling rate as such that, by running our data set level experiment, on the pseudo-labels generated using the above sampling rate, we will reach full-supervision results. For this reason, we ran the above experiment with various per image supervision (i.e., the percentage of labeled pixels).

We saw that, on average, labeling up to $0.05\%$ of the pixels in each training image is sufficient to reach full-supervision results, when using ImageNet pre-trained networks. This conclusion also holds for the low-data regime where only a subset of training images is annotated (e.g., 10\% of training data), meaning that it enables good representations of the label image.


For comparison, Bellver {\em et al.}~\cite{bellver2019budget} show similar behavior for the impact of using a subset of the training data while utilizing additional data sets. We reach preferable results with less supervision and without using additional data sets.
Our result shows that it is possible to reach full-supervision results using two orders of magnitudes less supervision than used by scribbles-based approaches \cite{Lin2016ScribbleSupSC, tang2018normalized, tang2018regularized, wang2019boundary}.

\section{Conclusions}

We have shown that an Oracle can train a semantic segmentation network from scratch, on a single image, using less than $0.1\%$ of labeled pixels. A network trained this way achieves more than $98\%$ accuracy when labeling the rest of the pixels in that particular image.

The Oracle uses Active Learning to train a segmentation network, which gives an upper bound on the amount of annotated pixels required for the task. It would be interesting to see if future algorithms will go even lower and achieve similar performance with even less labeled pixels.

To annotate an entire data set, an Oracle can train such networks for each of the training set images. These networks can then label the rest of the pixels in the images they were trained on. A semantic segmentation network trained on less than $0.03\%$ of labeled pixels, can achieve full-supervision results. This also holds for a low-data regime, where not all images are annotated.


These large scale experiments were designed to answer the question: how low can you go? In addition, they help clarify the contribution of various components.

For example, we found that initializing a network with random weights is sufficient if the Oracle is to train it for single image semantic segmentation. At the data set level, it is better to use an ImageNet pre-trained network for initialization.

The experiments also show clearly that order matters. Training a network with the correct order of annotated pixels leads to excellent results. Surprisingly, training with the correct order works even better than training on all the annotated pixels without order. Training with a random, or a reverse order hurts accuracy considerably.

Using an Oracle in the loop can help the future development of better annotation tools. For example, it can help us guide users where to annotate, or it can be used to evaluate future and better, Active Learning algorithms. Moving forward, we believe that those type of experiments offers a principled way to evaluate progress in the field and help us go even lower.

{\small
\bibliographystyle{ieee_fullname}
\bibliography{main}
}

\clearpage
{
\large\bf{Supplementary}%
\vspace*{12pt}%
\it%
}

We report the results of using classic sampling techniques, instead of the Active Learning approach reported in the main paper. We add another experiment that emphasizes the importance of the selected pixels across different architectures. We also introduce OracleNet, which predicts, at test time, the pixels to sample without having access to the ground truth. Moreover, we report the results of our experiments on the ADE20K data set. Lastly, we add a visualization of some of the results to demonstrate the power of our approach.

\section{Classic Sampling Methods}

The paper reports the results of an Active Learning approach to annotate an image. Here we report the results of several "classic" methods.



\paragraph{Classic Sampling Methods and Results}
We considered a couple of heuristics to sample points:
\begin{enumerate}
    \item {\bf Random} Randomly select $K$ pixels.
    \item {\bf Uniform} Split the image into a grid with $K$ tiles and sample a pixel randomly from each tile.
    \item {\bf Edge} Find all edge pixels in the label image and sample $K$ pixels.
    \item {\bf SLIC} Use SLIC~\cite{achanta2012slic} to split the image into $K$ segments and sample a pixel randomly from each segment.
    \item {\bf Geodesic} Use geodesic distance~\cite{criminisi2008geos} to sample $K$ pixels. We iteratively select the most distant pixel according to the geodesic distances on the currently selected pixels.
    \item {\bf PixelLabeling} Use Active Learning. We iteratively sample misclassified pixels corresponding to the lowest IoU class and retrain the network on the sampled pixels.
\end{enumerate}

\begin{table}[h]
\begin{center}
\begin{tabular}{l c}
Method & mIoU {$\pm$} Variance[\%]\\
\hline
Random  & {$80.47 \pm 14.31$} \\
Uniform & {$83.46 \pm 13.14$} \\
Edge    & {$42.62 \pm 18.35$} \\
SLIC    & {$84.12 \pm 12.22$} \\
Geodesic  & {$80.14 \pm 12.81$} \\
\textbf{PixelLabeling}  & $\mathbf{99.76 \pm 0.79}$ \\
\hline   
\end{tabular}
\end{center}
\caption{Image-Level Results. Pascal-VOC train set results (mIoU ± variance), using different sampling methods, for a sampling rate of 0.1\%. For each image, and for each sampling method, we train a network from scratch only on the labels given by the sampling method.}
\label{tab:different_sampling_methods_classic}
\end{table}

We tried each sampling technique with a sampling rate of $0.1\%$ and reported the results in Table~\ref{tab:different_sampling_methods_classic}.

We draw a number of observations from this experiment. First, we observe that {\bf Edges} gives the worst results with a mIoU of $42.62\%$ and quite a large variance of $18.35$. The second best method is {\bf SLIC} with a mIOU of $84.12\%$, but again with a very large variance. The best results are obtained by {\bf PixelLabeling} with mIOU of $99.76\%$ and fairly low variance. We also tried mixing the methods to increase spatial distribution (i.e., 'Edge' with 'Uniform'), but it did not produce better results.

We can also note that, similar to ~\cite{BharathICCV2011}, sampling all of the edges in the image (from both sides of the edge) without limiting the amount of sampled pixels, will indeed produce good results, but will increase the required number of sampled points by almost two orders of magnitude.

We conclude that heuristics are inferior to the Active Learning PixelLabeling algorithm presented in the main paper.


\section{OracleNet}
As we showed in our work, we use the ground truth to estimate the minimal subset of pixels that needs to be annotated such that we are able to reconstruct the entire label image only from the annotated pixels. Can this be extended to images for which we do not know their label map?

To do so, we replace the PixelLabeling algorithm with a network, termed {\em OracleNet} that will determine which pixels should be used. We treat the problem as a binary classification problem. The goal of OracleNet is to determine, for every pixel, if we should request its ground truth label. This is done by predicting a binary mask where the "on" pixels are the pixels that should be sampled. Crucially, OracleNet creates a binary mask map using the input RGB image only, without any access to the ground truth segmentation map.

This approach does not take the order of the pixels into account, but rather focuses, instead, on predicting the binary mask without any order. We prefer the simplicity of treating the binary mask without order and leaving a sequential pixel selection to future research.

OracleNet is based on the architecture of DeepLabV3+, but for binary segmentation. The training set consists of RGB images and their corresponding sampled pixels created using our PixelLabeling algorithm.  We train OracleNet on the Pascal VOC train set using 1200 image pairs and use 264 images for evaluation. Since only $0.1\%$ of the pixels in the ground truth binary segmentation maps are "on", the dataset is very imbalanced. To battle that, we use morphological operations to dilate (i.e., "inflate") the "on" pixels and use weighted Cross-Entropy loss with 0.8 weight on the "on" pixels compared to 0.2 on the "off" pixels. OracleNet is trained to predict two channels. One is the probability of a foreground pixel (i.e., a pixel that should be sampled), and another is the probability of a background pixel (i.e., a pixel that should not be sampled). 

After training OracleNet we run the following post-processing step. We iteratively perform weighted random sampling on the foreground probability channel to select the next pixel that should be sampled. We then zero out a $7 \times 7$ window around the selected pixel and repeat until enough pixels have been selected.

\newcommand{\mySizeForRandomVsSelectNet}{0.14}

\begin{figure*}[t]

\centering
\label{fig:examples_all}
\begin{tabular}{c c c c c c}
\multirow{2}{*}{\textbf{Image}}  & \multirow{2}{*}{\textbf{GT}} & \multirow{2}{*}{\textbf{Random}}  & \multirow{2}{*}{\textbf{Random}} & \multirow{2}{*}{\textbf{OracleNet}} & \multirow{2}{*}{\textbf{OracleNet}}\\
& & & & & \\
 & & \textbf{Points} & \textbf{Prediction} & \textbf{Points} & \textbf{Prediction} \\

\includegraphics[width=\mySizeForRandomVsSelectNet\linewidth]{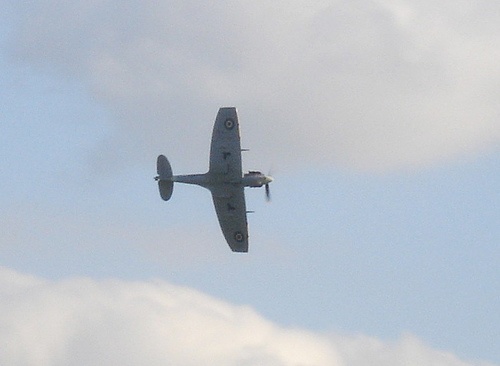} &
\includegraphics[width=\mySizeForRandomVsSelectNet\linewidth]{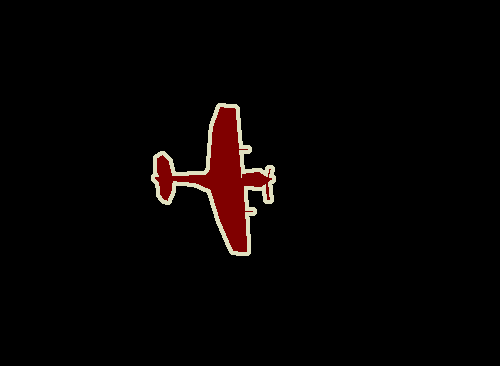} &
\includegraphics[width=\mySizeForRandomVsSelectNet\linewidth]{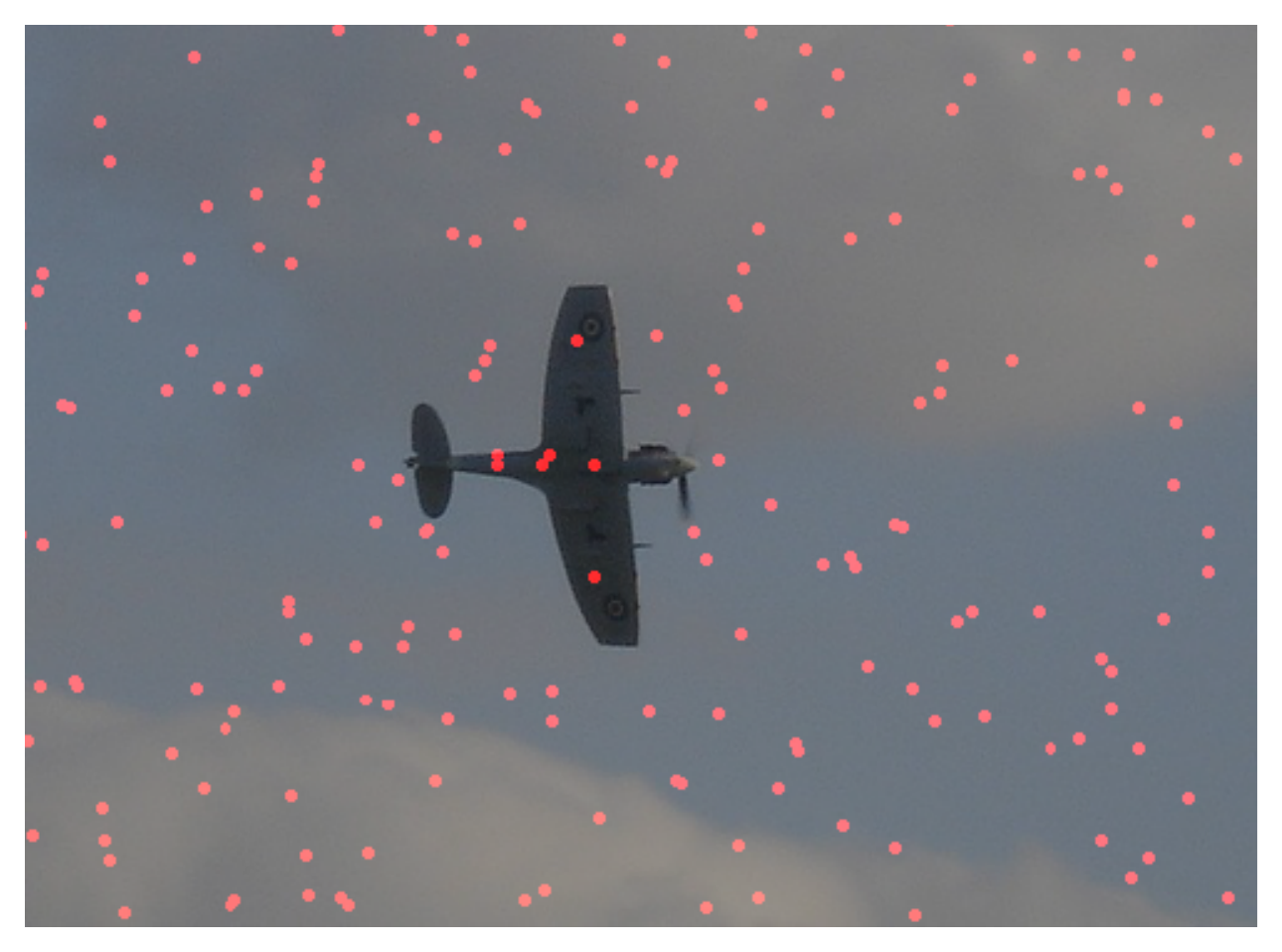} &
\includegraphics[width=\mySizeForRandomVsSelectNet\linewidth]{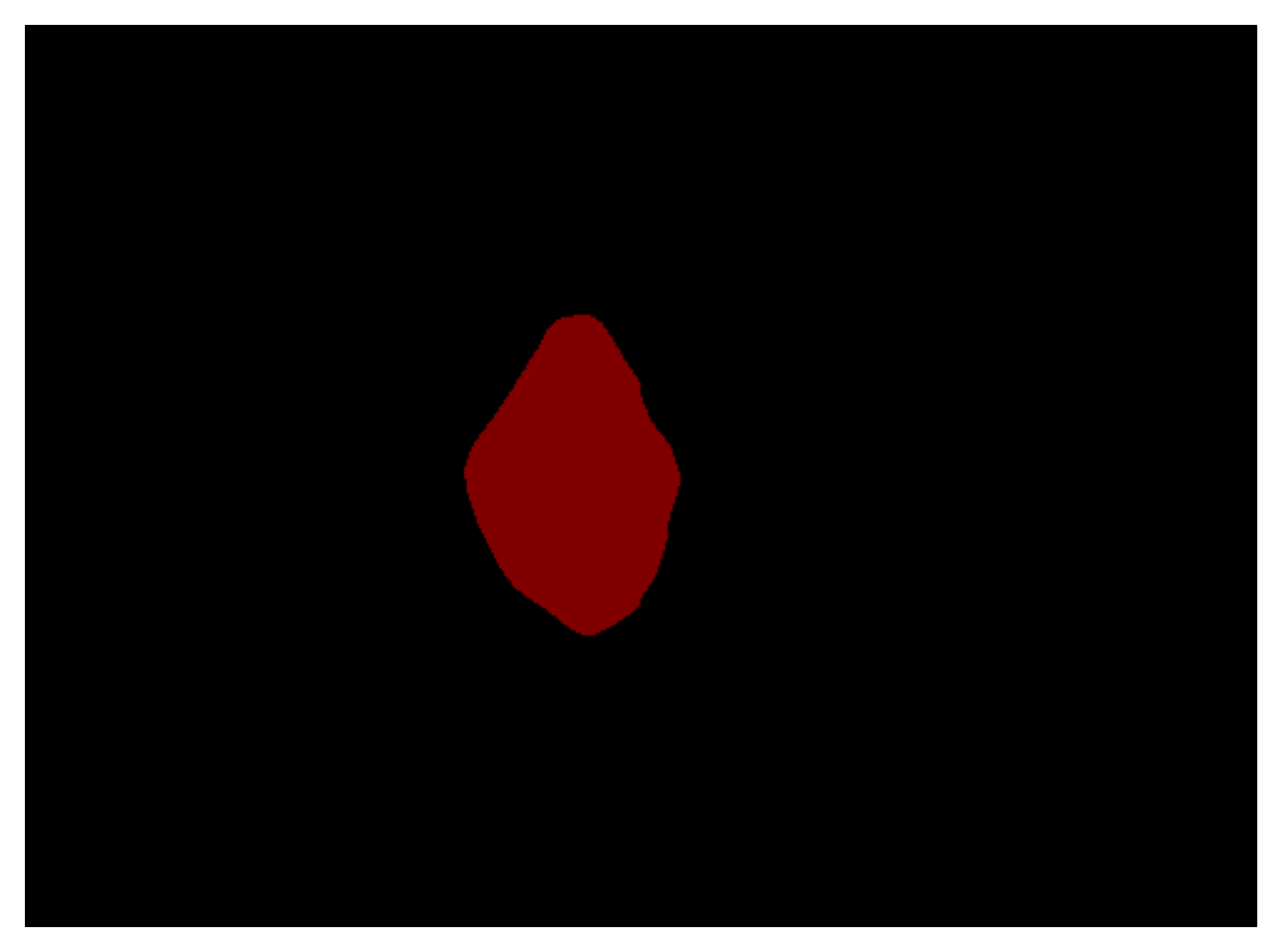} &
\includegraphics[width=\mySizeForRandomVsSelectNet\linewidth]{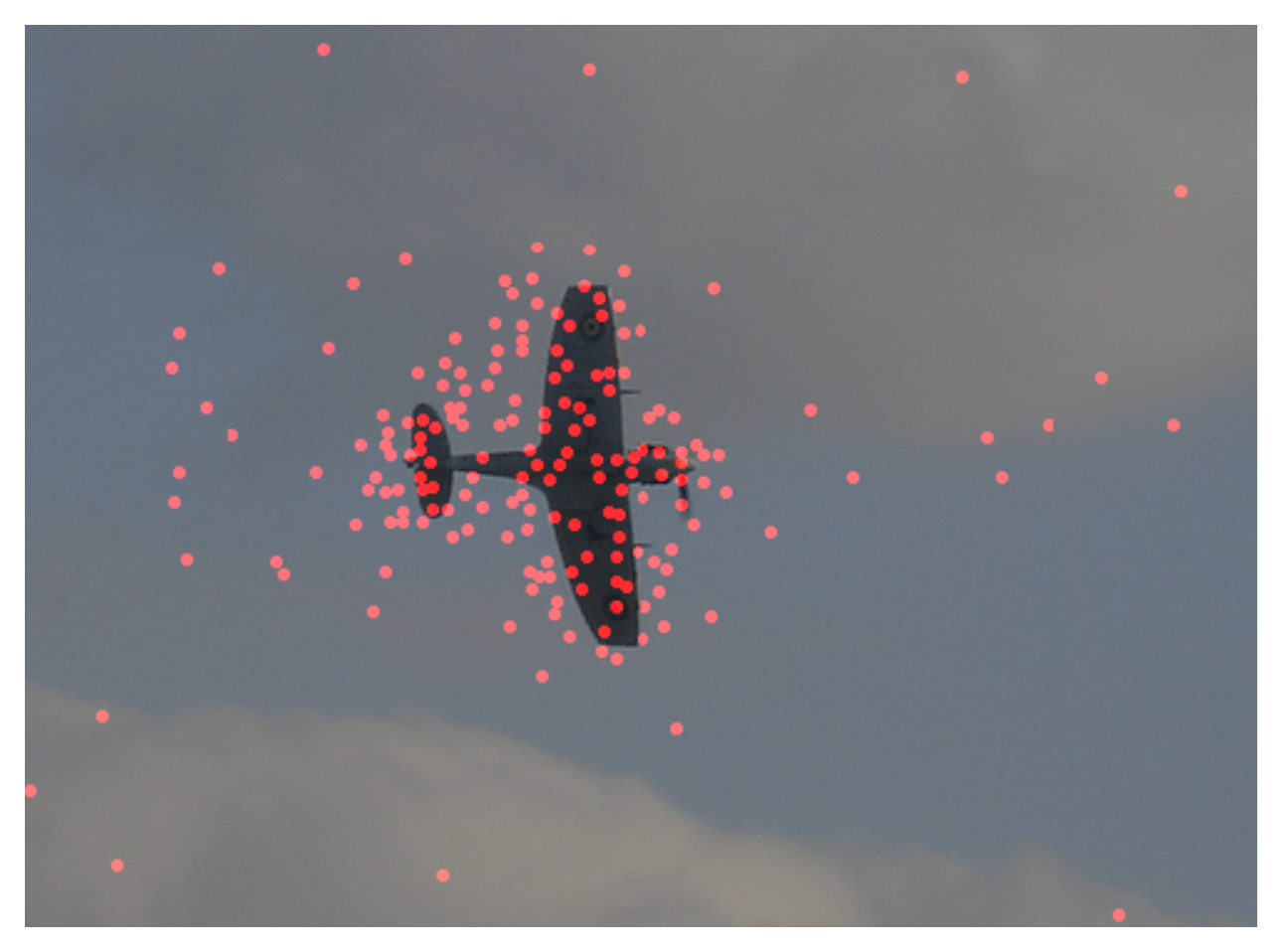} &
\includegraphics[width=\mySizeForRandomVsSelectNet\linewidth]{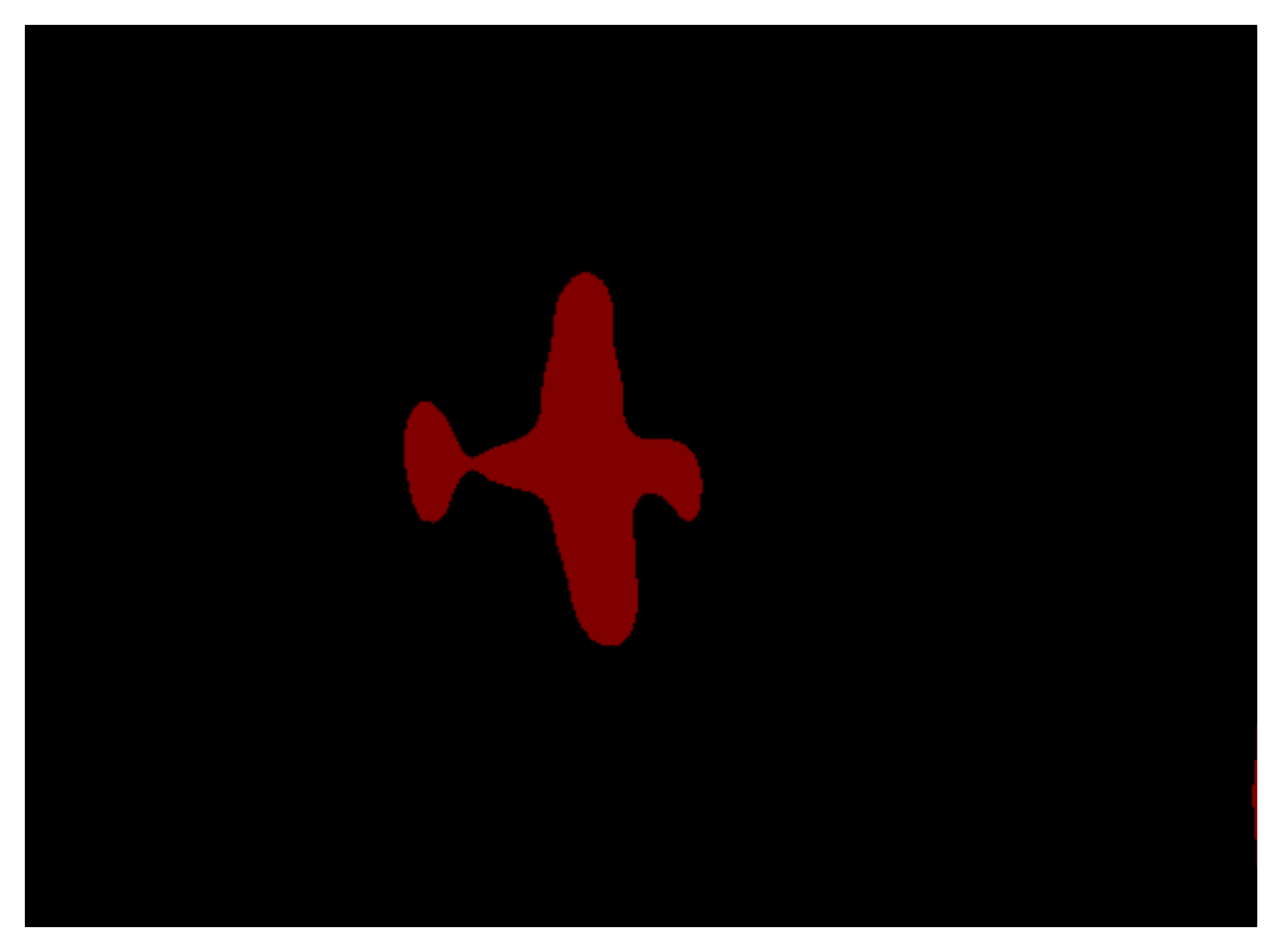}\\

\includegraphics[width=\mySizeForRandomVsSelectNet\linewidth]{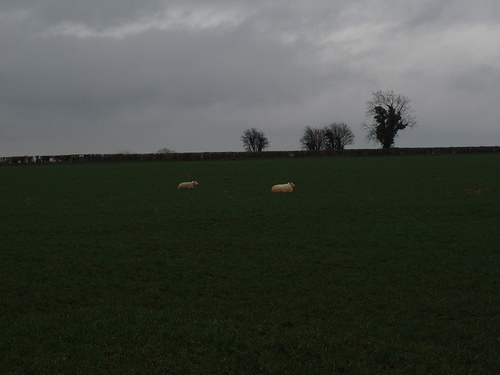} &
\includegraphics[width=\mySizeForRandomVsSelectNet\linewidth]{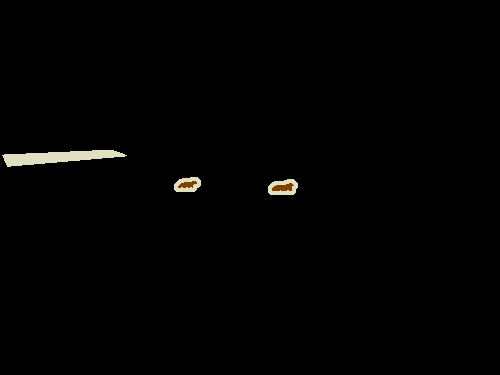} &
\includegraphics[width=\mySizeForRandomVsSelectNet\linewidth]{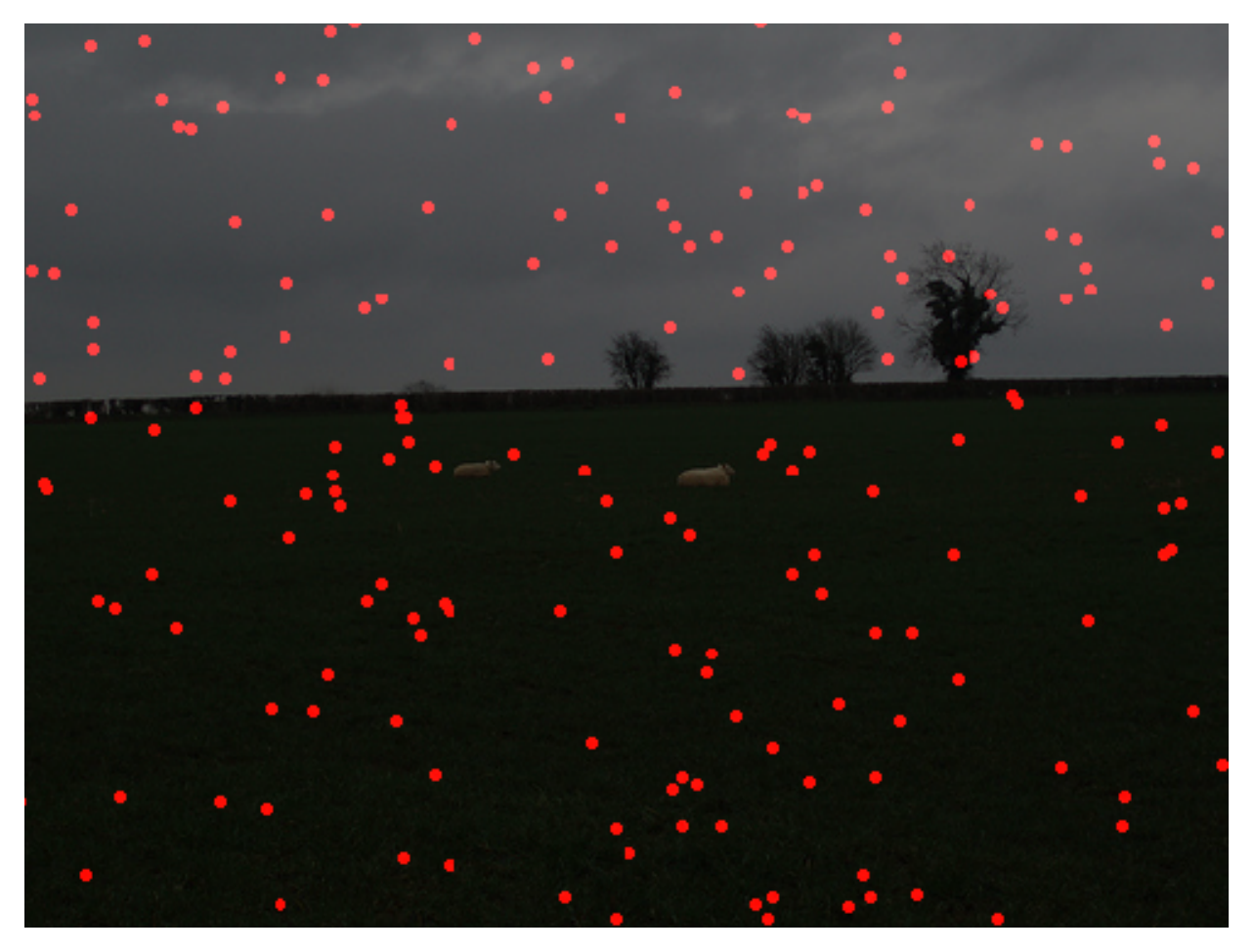} &
\includegraphics[width=\mySizeForRandomVsSelectNet\linewidth]{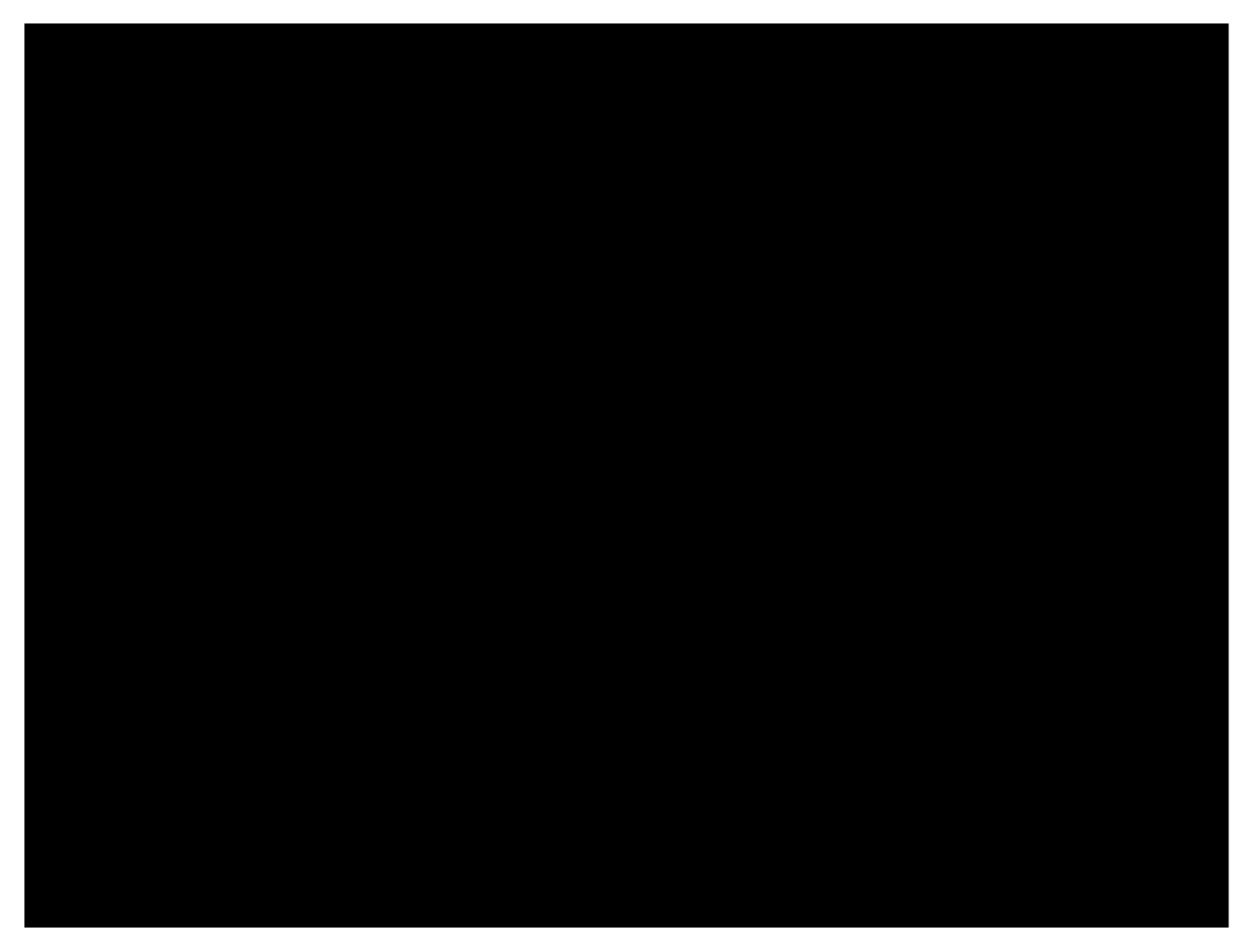} &
\includegraphics[width=\mySizeForRandomVsSelectNet\linewidth]{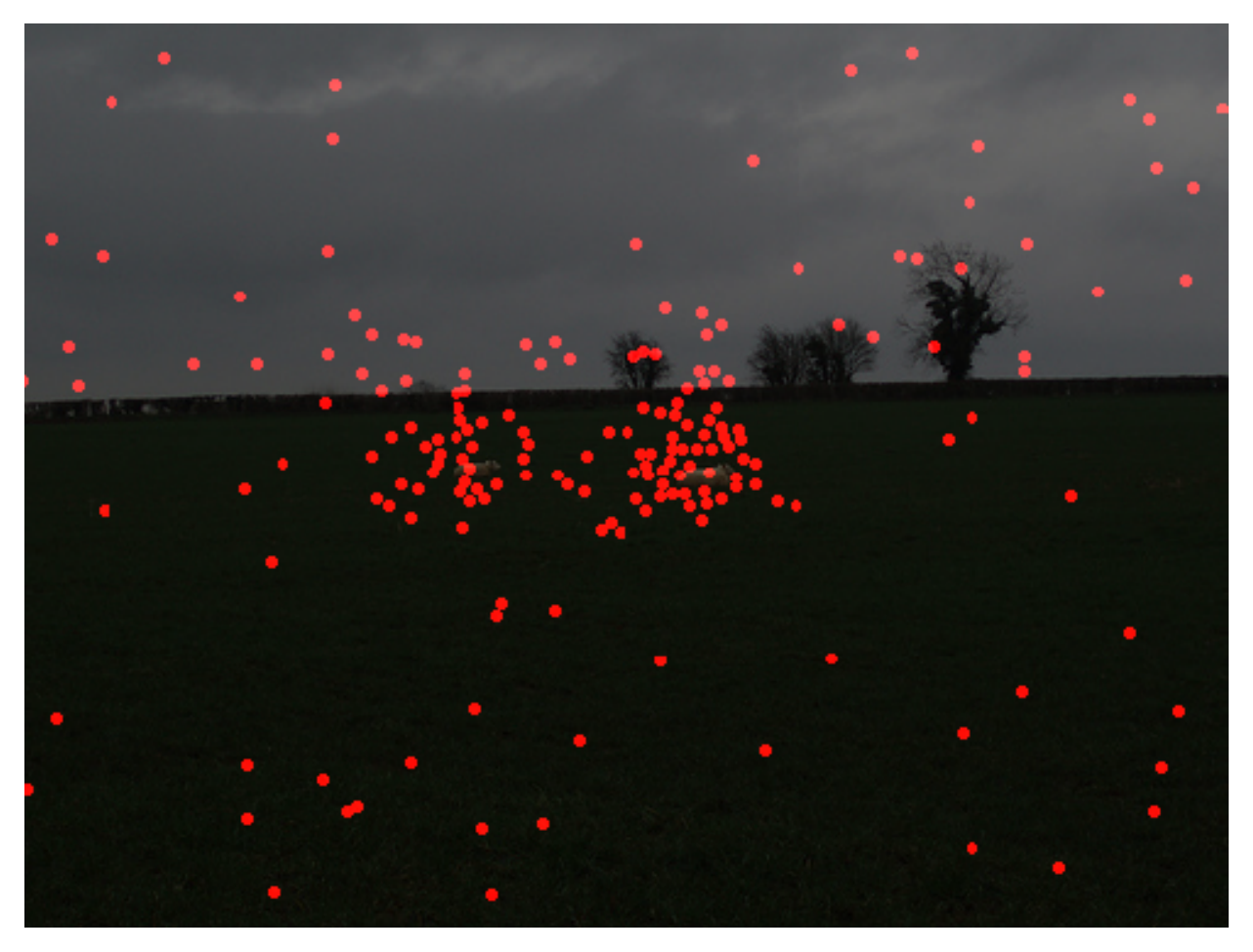} &
\includegraphics[width=\mySizeForRandomVsSelectNet\linewidth]{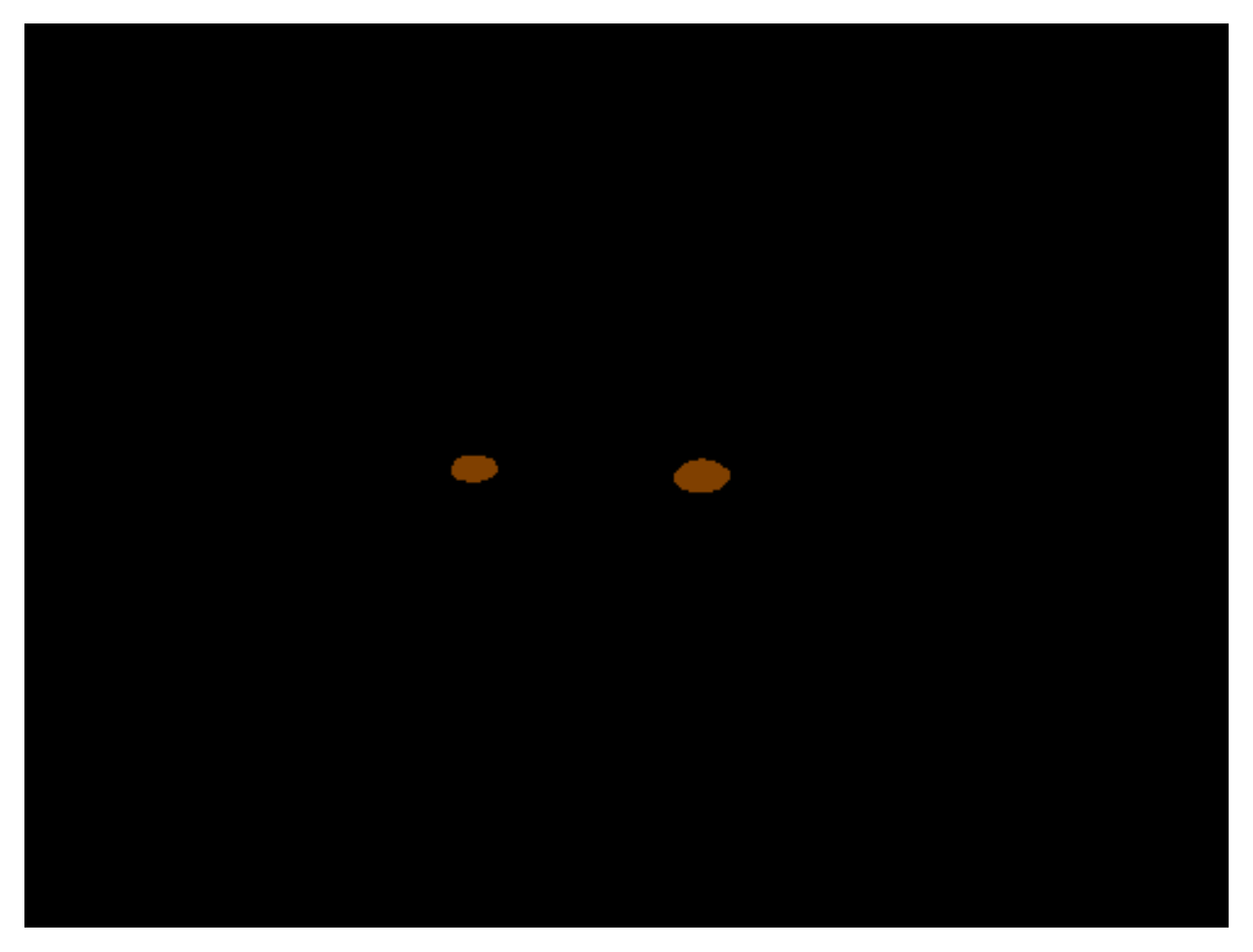}\\

\end{tabular}
\caption{Comparison between random sampling and OracleNet points and the corresponding resulting prediction.}
\label{fig:random_sampling_compared_to_network_points}
\end{figure*}

We compare three strategies to create the binary sampling mask. The first is based on random sampling, the second is based on our OracleNet, and the third is based on the PixelLabeling algorithm. Results are reported in Table~\ref{tab:single_image_result}. As can be seen, random sampling performs the worst, followed by OracleNet, which approaches the performance of the PixelLabeling algorithm. 

The PixelLabeling approach is based on an Oracle and can therefore be considered as an upper bound of any algorithm for single image semantic segmentation. This suggests that OracleNet is a viable approach to quickly label an entire dataset. Some examples can be seen in Figure~\ref{fig:random_sampling_compared_to_network_points}. It shows the pixels selected by our proposed OracleNet, as well as random sampling and PixelLabeling algorithm, and the single image semantic segmentation result. OracleNet clearly outperforms random sampling and scores close to the PixelLabeling result. We evaluated several other sampling strategies, but they all failed to achieve the performance of OracleNet.

\begin{figure*}[]

\centering
\begin{tabular}{c c c c c}
\multirow{2}{*}{\textbf{Image}}  & \multirow{2}{*}{\textbf{GT}} & \textbf{Points}  & \textbf{OracleNet} & \textbf{PixelLabeling}\\
&             &             &             \textbf{Prediction}   & \textbf{Prediction} \\
\includegraphics[width=0.16\linewidth]{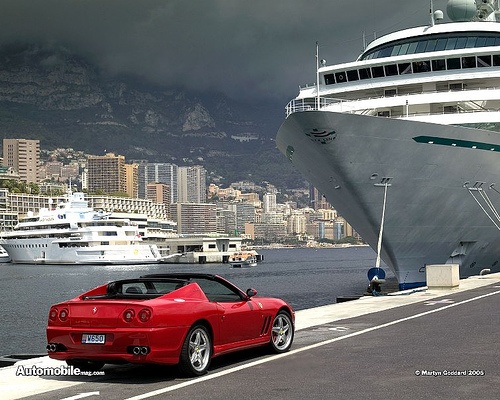} & 
\includegraphics[width=0.16\linewidth]{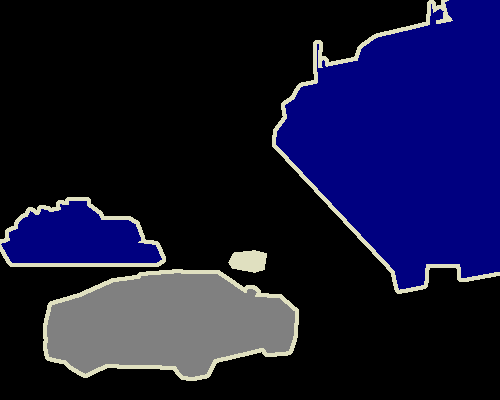} & 
\includegraphics[width=0.16\linewidth]{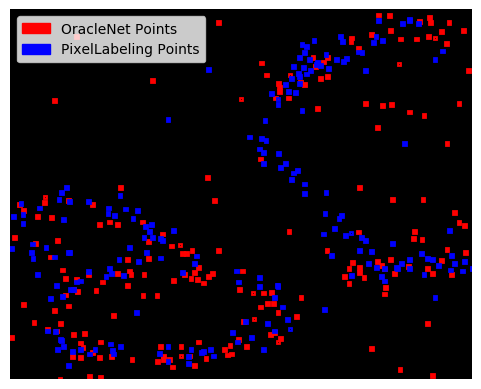} & 
\includegraphics[width=0.16\linewidth]{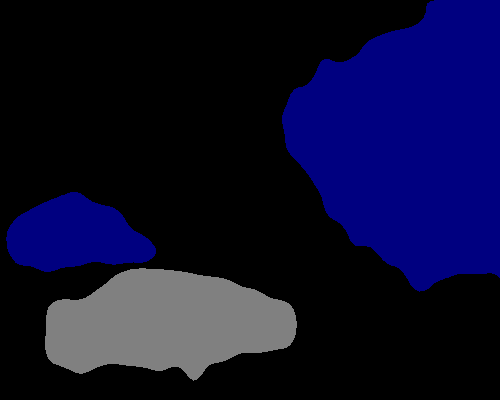} & 
\includegraphics[width=0.16\linewidth]{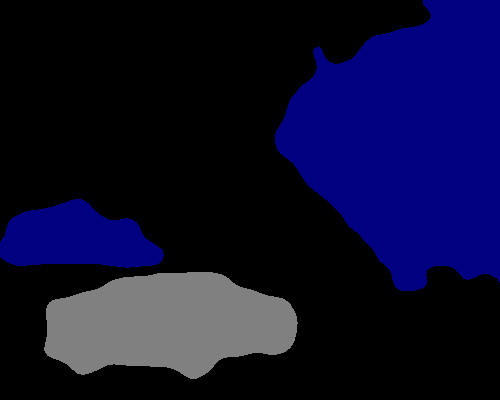} \\

\end{tabular}
\caption{Visualization of an input image, corresponding ground truth, the selected points using the OracleNet and the PixelLabeling algorithm and the prediction result when training a network on the points. OracleNet prediction IOU score 97.2\% compared to 99.5\% with the PixelLabeling points.}
\label{fig:pseudo_optimal_compare_to_network_all}

\end{figure*}

The performance of OracleNet can be understood by comparing the pixels it selected against those selected by PixelLabeling algorithm. An example is shown in Figure~\ref{fig:pseudo_optimal_compare_to_network_all}. As can be seen, the selected pixels are distributed in a similar fashion, leading to similar single image segmentation results.

\begin{table}
\begin{center}
\begin{tabular}{l c}
Method & mIoU Score[\%]\\
\hline
PixelLabeling & 96.7\% \\
OracleNet & 93.6\% \\
Random & 86.3\% \\
\hline   
\end{tabular}
\end{center}
\caption{Single Image Semantic Segmentation: mIoU score of different sampling methods. Results are reported on a subset of 264 images from Pascal VOC train set. OracleNet is trained on 1200 images from the Pascal VOC train set and evaluated on the 264 other images in the train set.}
\label{tab:single_image_result}
\end{table}

\begin{table}
\begin{center}
\begin{tabular}{l c}
Method & mIoU Score[\%]\\
\hline
Fully-Supervised & 66.1\% \\
PixelLabeling & 65.7\% \\
OracleNet & 63.6\% \\
Random & 60.9\% \\
\hline   
\end{tabular}
\end{center}
\caption{Dataset level annotation: mIoU score of a trained network using different methodologies. The network was trained on 300 images from the Pascal VOC validation set and evaluated on the rest of the validation set.}
\label{tab:end_to_end_expr}
\end{table}

In another experiment, we show how to use our proposed OracleNet in order to annotate an entire dataset. To this end, we took 300 images from the Pascal VOC validation set, that the OracleNet was not trained on, as training, and the rest of the validation set images for testing. Using this train/test split, we trained a semantic segmentation network as follows:
\begin{enumerate}
    \item {\bf OracleNet} For each image, use OracleNet to estimate the binary sampling mask (0.1\%) and obtain labels for each selected pixel.
    \item {\bf Pseudo-Labels} For each image, train a segmentation network using only the given labels, to estimate the pseudo-labels for the entire image.
    \item {\bf Network Training} Train a segmentation model on those pseudo-labels and log the score on the test set.
\end{enumerate}


Table~\ref{tab:end_to_end_expr} shows the results of the experiment. As expected, PixelLabeling performs the best, with OracleNet coming a close second. The random sampling approach gives the worst results.

\section{Points Importance Across Architectures}

\begin{figure}[h!]
\begin{center}
\begin{tabular}{c c}
\textbf{Correct Order} & \textbf{No Order}  \\
\includegraphics[width=0.45\linewidth]{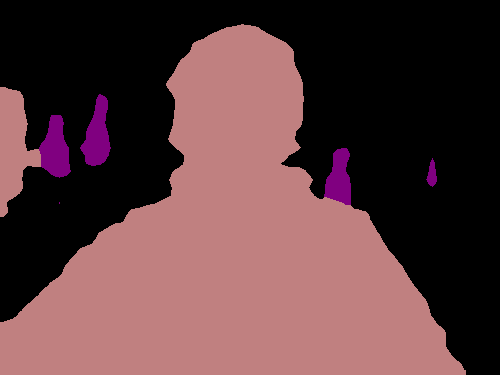} &
\includegraphics[width=0.45\linewidth]{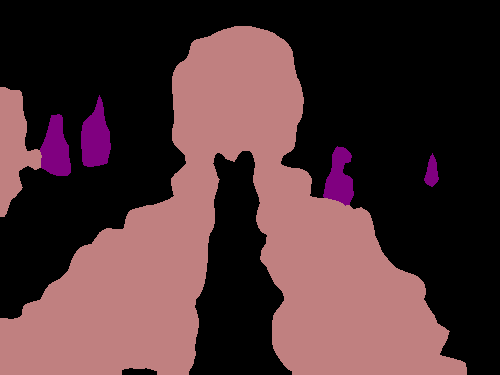}\\
\end{tabular}
\caption{Example of PSPNet model prediction using points obtained with DeepLabV3+ model. We show a comparison between training in the \emph{correct} order to training on all the points without order. Results are similar to what is shown in Figure 5 in the paper.}
\label{fig:order_on_different_architecture}
\end{center}
\end{figure}

We conducted an additional experiment to further \emph{emphasize} the importance of selecting the most \emph{informative} pixels, to add on top of the other experiments. We aimed to check whether the selected points are important for our specific model architecture, or rather they are transferable across different models. 
To answer this question, we took the selected points and their order (i.e., extracted using DeepLabV3+) and trained a new network architecture (PSPNet \cite{zhao2017pyramid}) using the extracted points, and logged the resulting mIoU.
Table~\ref{tab:order_on_different_architecture} shows the results over Pascal-VOC dataset, while Figure~\ref{fig:order_on_different_architecture} shows a specific visual example.
We can see that the order is important across architectures, which means that the extracted points are \emph{informative} across the different models. Furthermore, we see that keeping the correct training order is also important.
We believe that additional research can be done on this interesting phenomenon.
No optimization of hyper-parameters was done in our experiments (architecture, forward-backward iterations), which shows that the naive implementation gains large improvement of existing methods.

\begin{table}[h!]
\begin{center}
\begin{tabular}{l c c}
Method & DeepLab & PSPNet\\
\hline
No Order & {$94.79 \pm 5.92$} & {$82.36 \pm 14.29$}\\
\hline
Random Order & {$94.69 \pm 7.09$} & {$85.84 \pm 8.69$}\\
\hline
Inverse Order & {$81.06 \pm 12.52$} & {$77.73 \pm 10.82$}\\
\hline
\textbf{Correct Order} & $\mathbf{99.76 \pm 0.79}$ & $\mathbf{94.35 \pm 5.53}$ \\
\hline
\end{tabular}
\end{center}
\caption{We train a PSPNet model using points extracted using DeepLabV3+ model. We show the resulting ${(mIoU \pm variance)}$ for each training procedure, over Pascal-VOC.}
\label{tab:order_on_different_architecture}
\end{table}

\section{ADE20K Data Set}
The data set contains over 20000 training images and 2000 validation images, with 150 different classes. We used all the validation images for testing, and sampled $25\%$ of the training images (i.e., 5000 images) such that the selected images will have the same label distribution found in the entire training set.
As opposed to Pascal-VOC and Kvasir-SEG datasets, each image has a large number of classes (on average, 10 different classes per image), while having a similar image size. Also, the state-of-the-art network for this data set does not reach $50\%$ mIoU score~\cite{zhang2020resnest}, meaning that this is a challenging data set. As a result, we observed that we need to increase the sampling rate in order to achieve a good label prediction. This suggests that the sampling rate depends on the number of labels per image, and not just the image size.

\subsection{ADE20K Results}
We repeated the experiments discussed in the paper and reported the results. Because of limited computational resources, we tweaked some of the parameters of the algorithm to speed up the process. Specifically, we sampled $30$ pixels in each iteration (instead of $5$ used in the main paper), and as a result, increased the number of forward-backward iterations from $10$ to $20$. We also set the threshold parameter to $97\%$, as the images contain a larger number of classes, which hurts the mIoU measurement. We believe that sampling fewer pixels in each iteration will improve results.



\begin{figure}[htb!]
\begin{center}
\begin{tabular}{c}
\includegraphics[width=0.95\linewidth]{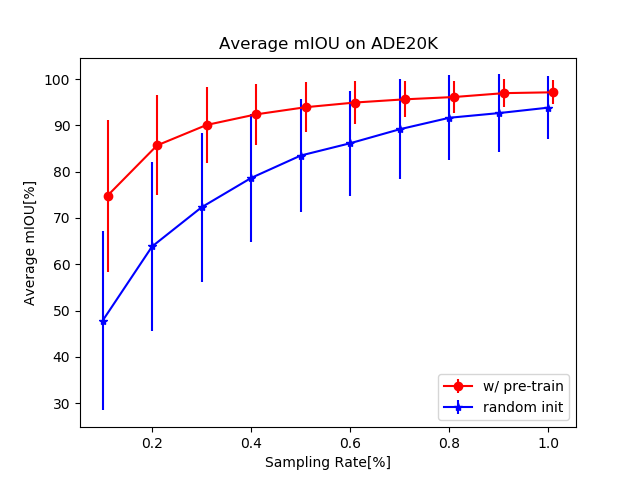}\\
\end{tabular}
\caption{Image-Level visualization - Resulting mIoU scores and variance (bars), for different sampling rates, obtained using the proposed algorithm, and averaged across 25\% of ADE20K training set. We compare a randomly initialized network results to an ImageNet pre-trained one.}
\label{fig:with_and_without_imagenet_ADE}
\end{center}
\end{figure}

We used a sampling rate of $1\%$ and ran our experiment, for a randomly initialized network as well as an ImageNet pre-trained one. The results are shown in Figure~\ref{fig:with_and_without_imagenet_ADE}. We see that effectively, due to the threshold value (i.e., maximal mIoU score that stops sampling more points), an average sampling rate of $0.5\%$ yields $95.5\%$ mIoU score, which, according to the data set level experiment, is enough to reach near full-supervision results.

As before, starting from a model pre-trained on ImageNet converges faster than using a randomly initialized network. However, when increasing the sampling rate, the gap shrinks.

\begin{table}[h]
\centering
\begin{tabular}{l | c |c }
\hline
\textbf{Method} & \textbf{Randomly Initialized} &\textbf{Pre-Trained}\\
        \hline
No Order & {$67.35 \pm 16.66$} & {$94.88 \pm 3.66$} \\
\hline
Random Order & {$82.05 \pm 11.44$} & {$91.78 \pm 4.73$} \\
\hline
Inverse Order & {$78.91 \pm 12.69$} & {$91.55 \pm 4.85$}\\
\hline
\textbf{Correct Order} & $\mathbf{93.83 \pm 6.85}$ & $\mathbf{97.14 \pm 2.56}$ \\
\hline
\multicolumn{2}{c}{} 
\end{tabular}\\
\caption{Distribution of mIoU results over 25\% of ADE20K training set. Results (mean $\pm$ variance) are shown for each different network training procedure using a pre-trained/randomly initialized network.}
\label{tab:method::order_of_training_ADE}
\end{table}

\paragraph{Order experiment:} We also repeated the experiment to see the importance of the training order. Results are reported in Table~\ref{tab:method::order_of_training_ADE}. Training in the \emph{Correct} order yields the best results, and by a large gap, for a randomly initialized network. Training on the entire available labels without any order (i.e., 'No Order') is a good option for the pre-trained network, but hurts performance considerably for a randomly initialized network. We can also attribute the success of training without order (i.e., 'No Order') to the fact that we sampled a large number of pixels in each iteration, which in turn reduced the number of iterations.

\paragraph{Data set level experiment:} We repeated the data set experiment for the ADE20k data set. We used the annotations of the $25\%$ of labeled images and ignored the labels of the remaining training data (we created pseudo-labels for them, as explained in the main paper). We compared the performance of a network trained using all labeled pixels (i.e., for the $25\%$ of labeled images), with a network trained using $0.5\%$ of labeled pixels (corresponding to a sampling rate of $0.7\%$).
While training on all available labeled pixels resulted in a performance of $29\%$ mIoU, training on just 0.5\% of pixels label, resulted in a decrease of $0.3\%$, to a score of $28.7\%$.
For comparison, a fully-supervised model (i.e., with all training images) reached $39\%$.

It should be noted that we only ran this experiment using a pre-trained network. We conclude that even with a long-tailed distribution like ADE20K our experiment still shows promising results, using a sparse subset of annotated pixels.

\subsection{Visualizations}
Finally, we show some examples of our experiments. These examples demonstrate the importance of order and the quality of the prediction.

Figures~\ref{fig:results::different_order_inference_without_pretrain} and ~\ref{fig:results::different_order_inference_with_pretrain} show the results of a randomly initialized network and an ImageNet pre-trained network. These examples demonstrate the importance of the order. As can be seen, following the correct order yields visually pleasing results.

Given the correct order, the trained network produces good segmentation maps, as can be seen in Figures~\ref{fig:sampled_pts_visualization_without_pre_train} and~\ref{fig:sampled_pts_visualization_with_pre_train} that show results for randomly initialized and ImageNet pre-trained networks, respectively.

\newcommand{\mysizeWithoutTrain}{0.14}

\begin{figure*}[htb!]
\begin{center}
\begin{tabular}{c c c c c c}

\textbf{Image} & \textbf{Labels} & \textbf{Correct Order} & \textbf{No Order}  & \textbf{Random Order} & \textbf{Reverse Order}  \\
\includegraphics[width=\mysizeWithoutTrain\linewidth]{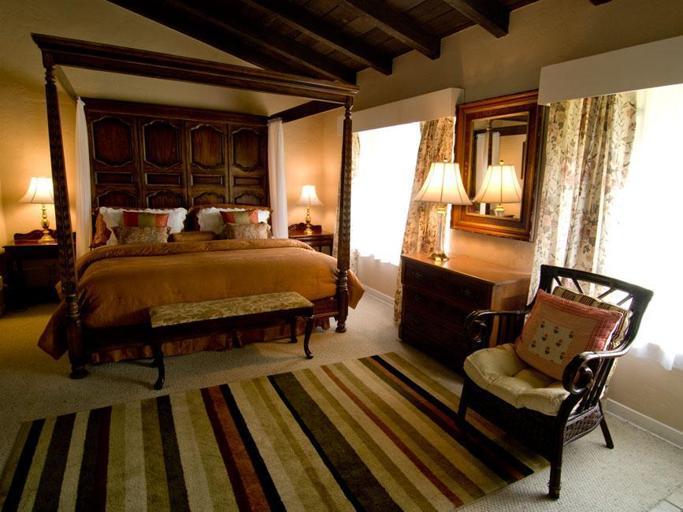} &
\includegraphics[width=\mysizeWithoutTrain\linewidth]{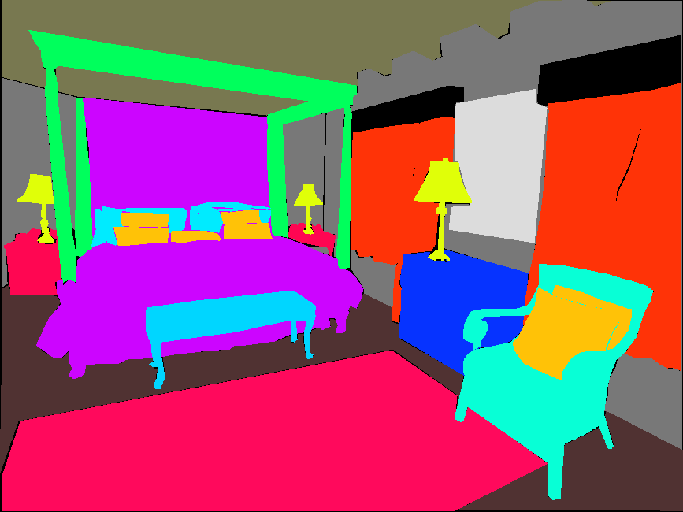} &
\includegraphics[width=\mysizeWithoutTrain\linewidth]{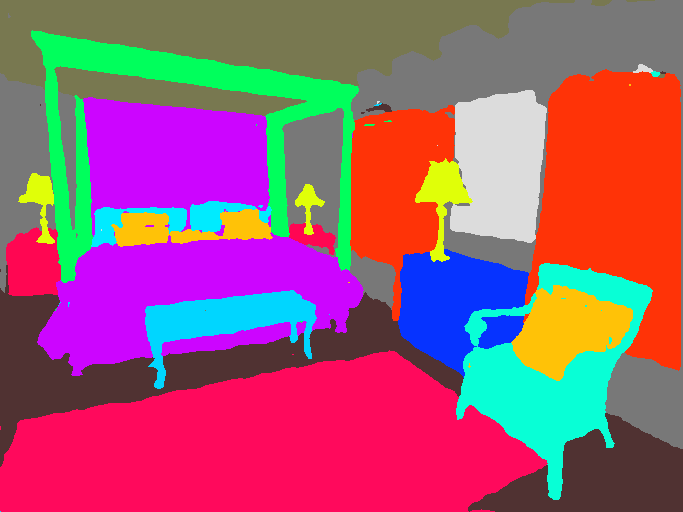} &
\includegraphics[width=\mysizeWithoutTrain\linewidth]{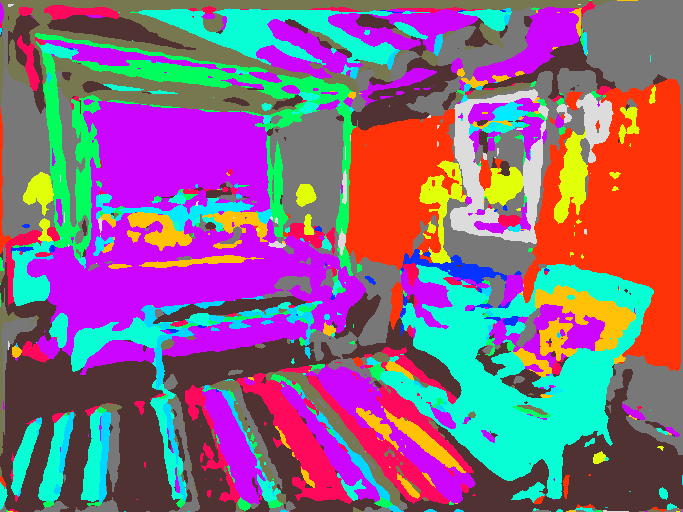} &
\includegraphics[width=\mysizeWithoutTrain\linewidth]{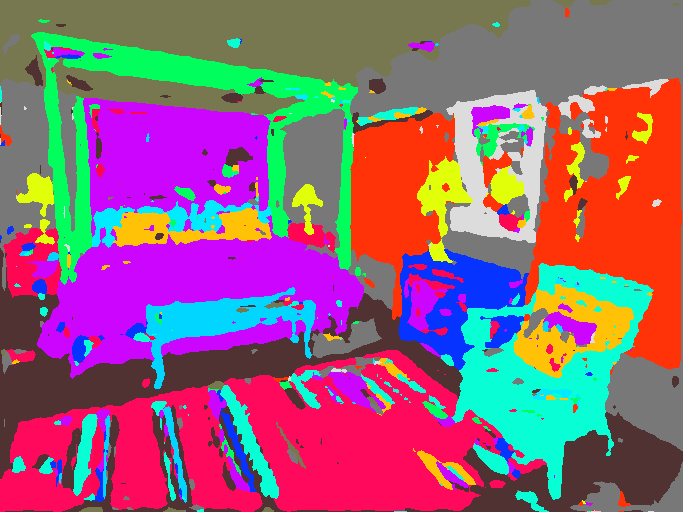} &
\includegraphics[width=\mysizeWithoutTrain\linewidth]{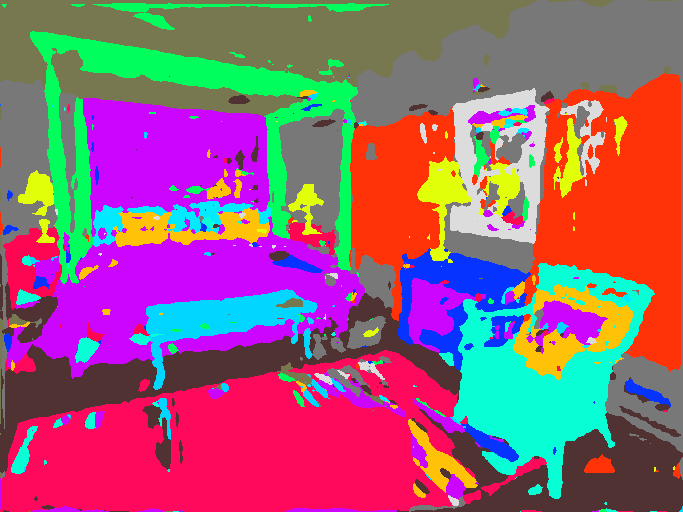} \\

\includegraphics[width=\mysizeWithoutTrain\linewidth]{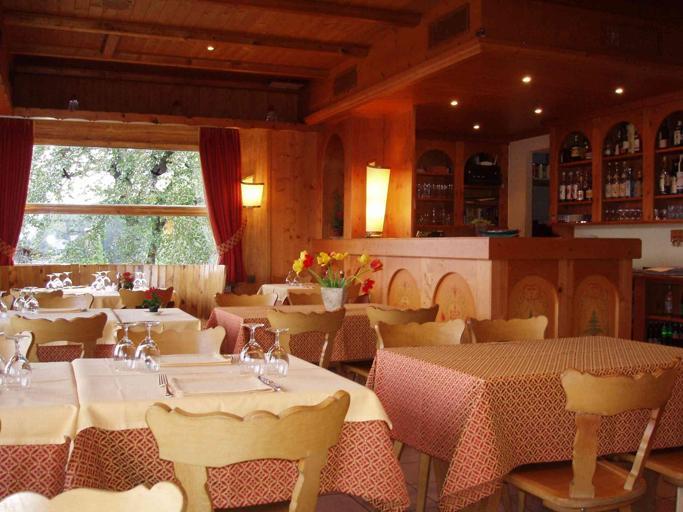} &
\includegraphics[width=\mysizeWithoutTrain\linewidth]{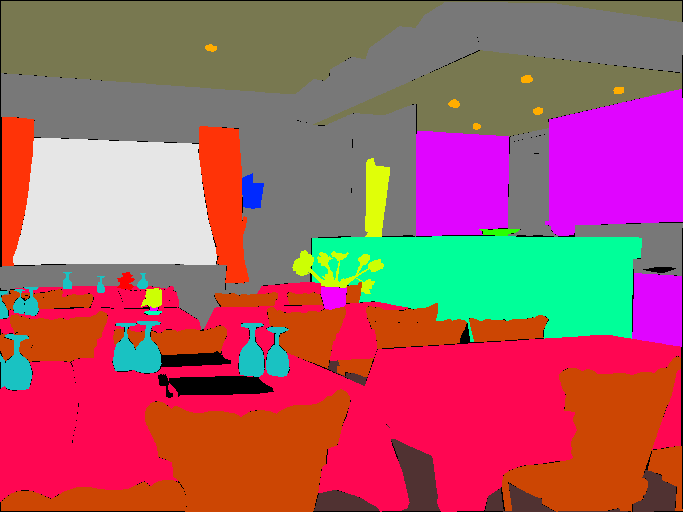} &
\includegraphics[width=\mysizeWithoutTrain\linewidth]{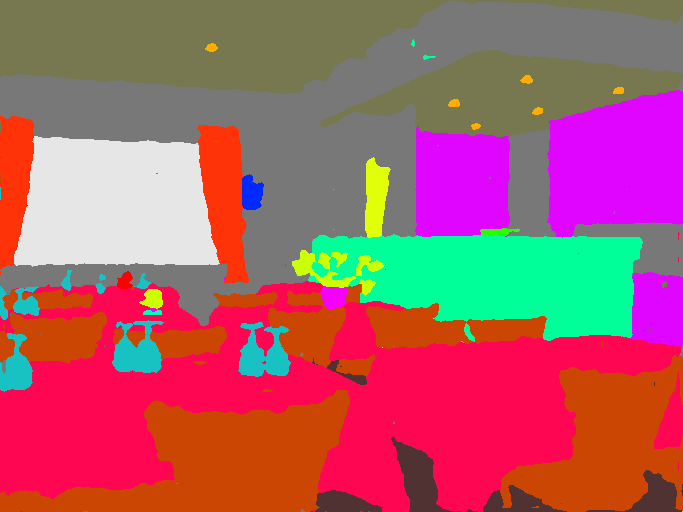} &
\includegraphics[width=\mysizeWithoutTrain\linewidth]{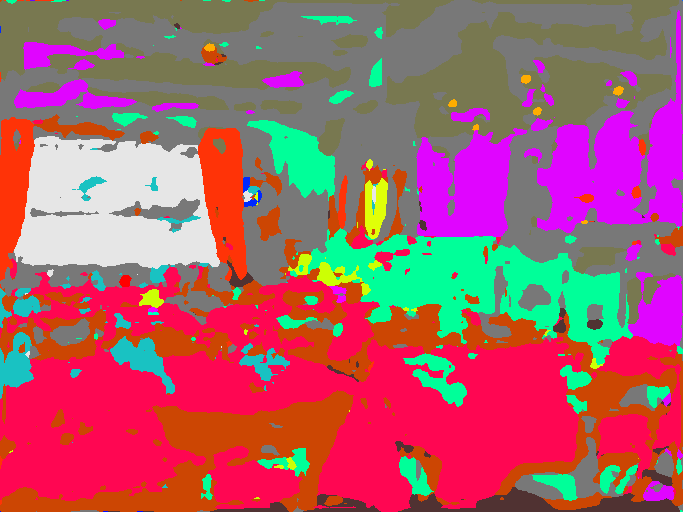} &
\includegraphics[width=\mysizeWithoutTrain\linewidth]{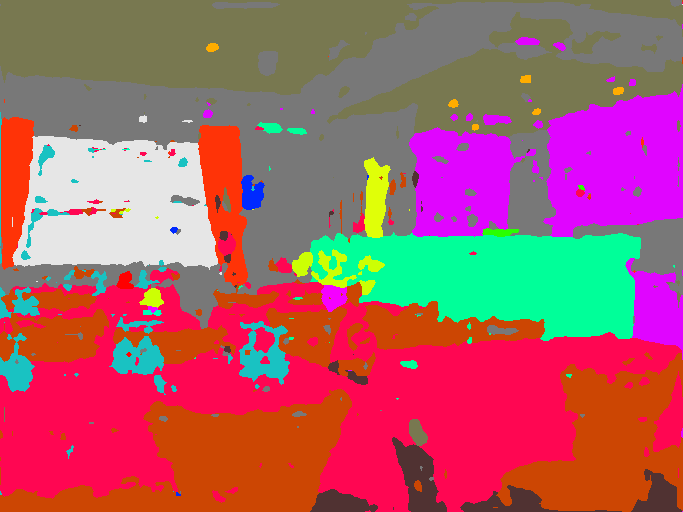} &
\includegraphics[width=\mysizeWithoutTrain\linewidth]{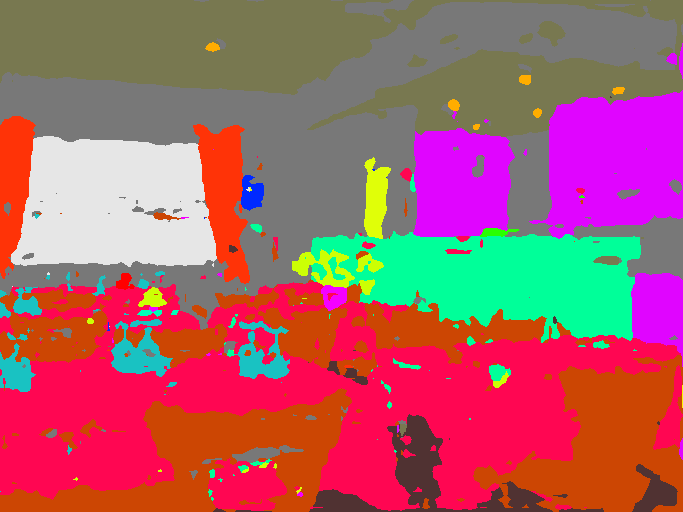} \\

\includegraphics[width=\mysizeWithoutTrain\linewidth]{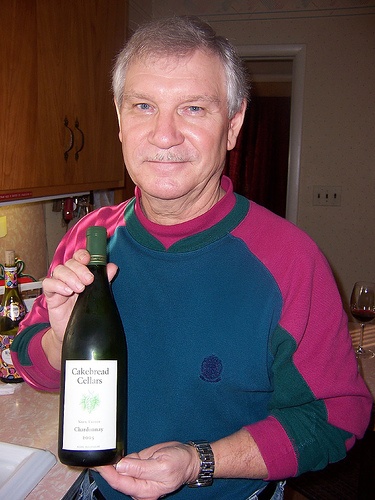} &
\includegraphics[width=\mysizeWithoutTrain\linewidth]{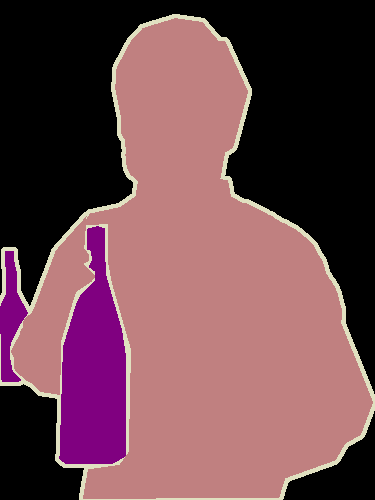} &
\includegraphics[width=\mysizeWithoutTrain\linewidth]{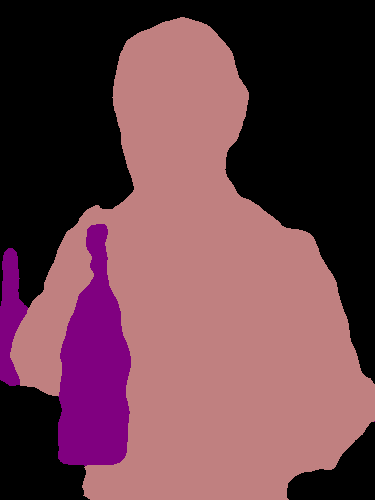} &
\includegraphics[width=\mysizeWithoutTrain\linewidth]{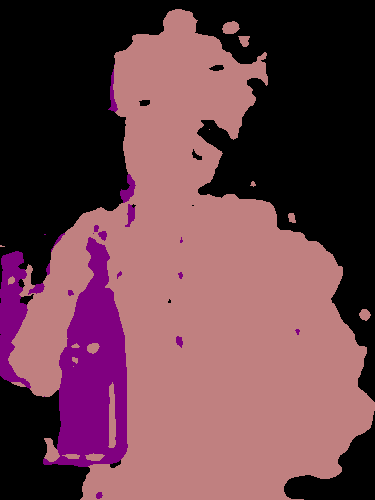} &
\includegraphics[width=\mysizeWithoutTrain\linewidth]{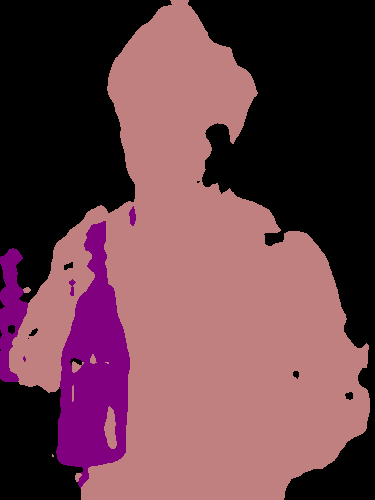} &
\includegraphics[width=\mysizeWithoutTrain\linewidth]{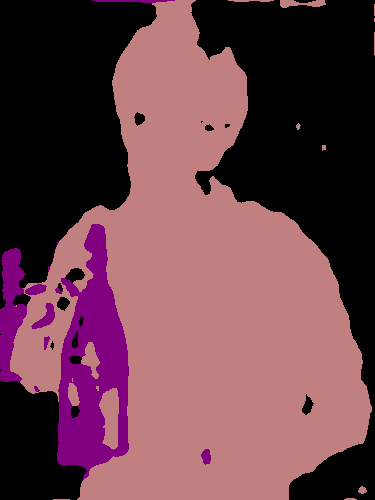} \\

\includegraphics[width=\mysizeWithoutTrain\linewidth]{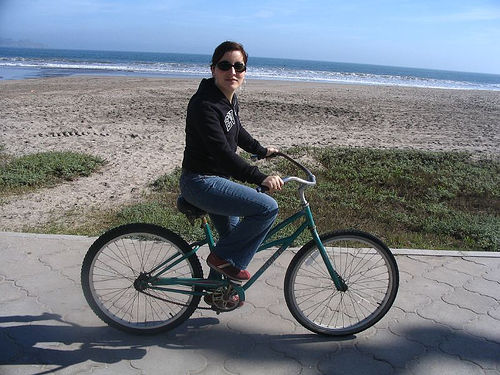} &
\includegraphics[width=\mysizeWithoutTrain\linewidth]{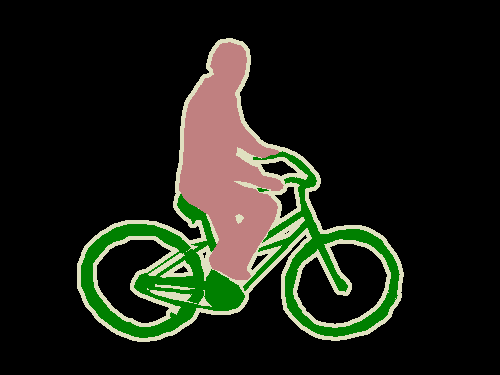} &
\includegraphics[width=\mysizeWithoutTrain\linewidth]{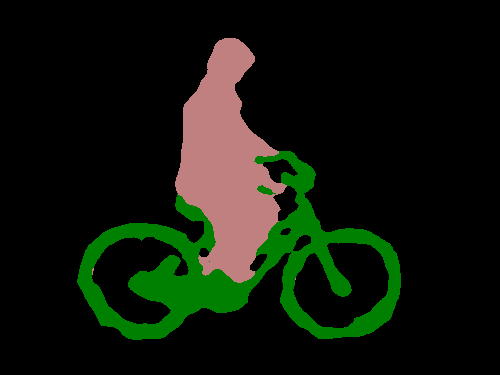} &
\includegraphics[width=\mysizeWithoutTrain\linewidth]{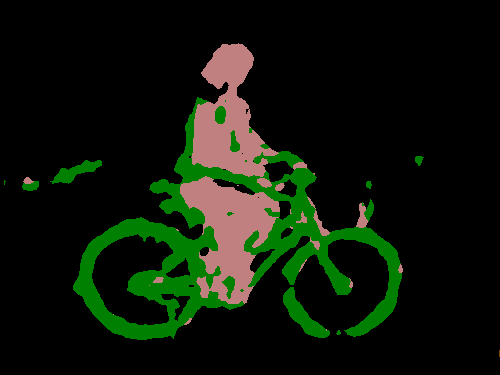} &
\includegraphics[width=\mysizeWithoutTrain\linewidth]{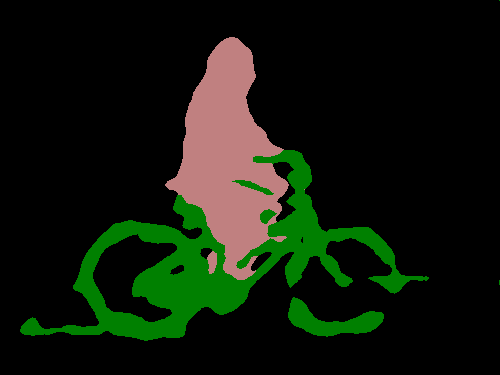} &
\includegraphics[width=\mysizeWithoutTrain\linewidth]{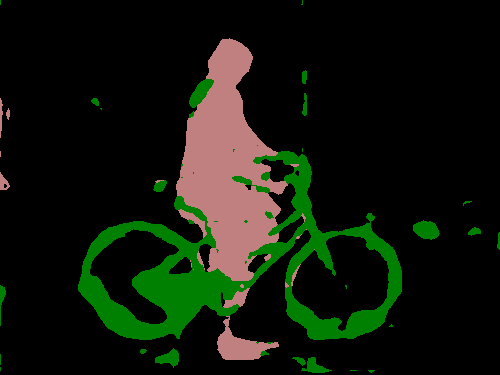} \\

\end{tabular}
\caption{{\bf Single-Image Order Training (random):} Results for different training techniques using networks with \textbf{randomly initialized} weights. Each image represents a different network prediction, using four different network training procedures over a sparse 0.1\% labeled image. In each data set, some pixels are marked with an ignore index, such that we do not back-propagate on those pixels.}
\label{fig:results::different_order_inference_without_pretrain}
\end{center}
\end{figure*}
\newcommand{\mysizeWith}{0.142}

\begin{figure*}[htb!]
\begin{center}
\begin{tabular}{c c c c c c}

\textbf{Image} & \textbf{Labels} & \textbf{Correct Order} & \textbf{No Order}  & \textbf{Random Order} & \textbf{Reverse Order}  \\
\includegraphics[width=\mysizeWith\linewidth]{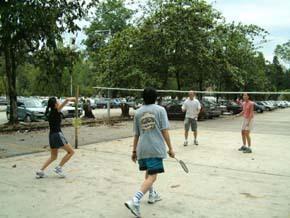} &
\includegraphics[width=\mysizeWith\linewidth]{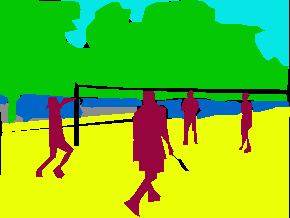} &
\includegraphics[width=\mysizeWith\linewidth]{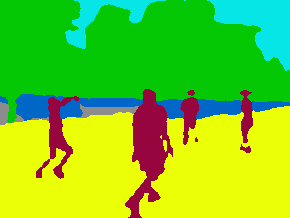} &
\includegraphics[width=\mysizeWith\linewidth]{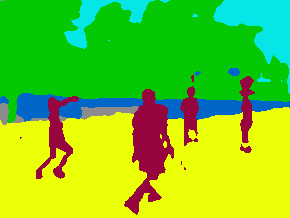} &
\includegraphics[width=\mysizeWith\linewidth]{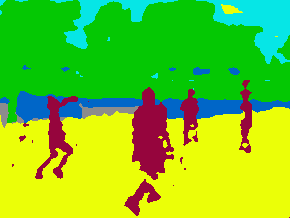} &
\includegraphics[width=\mysizeWith\linewidth]{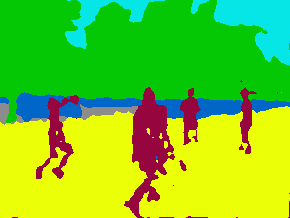} \\

\includegraphics[width=\mysizeWith\linewidth]{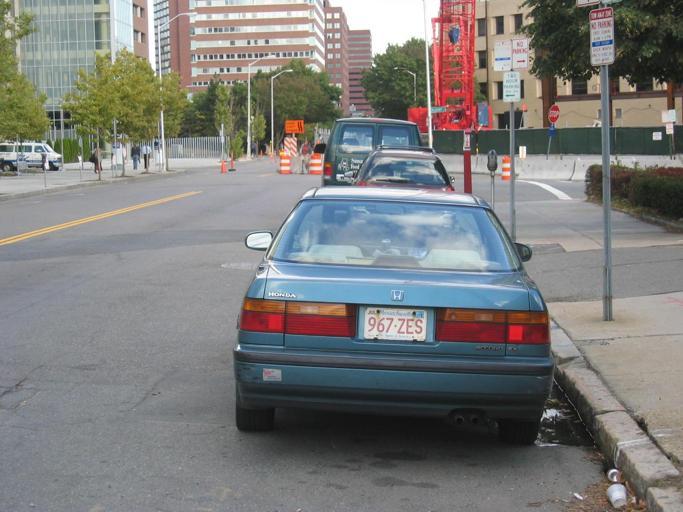} &
\includegraphics[width=\mysizeWith\linewidth]{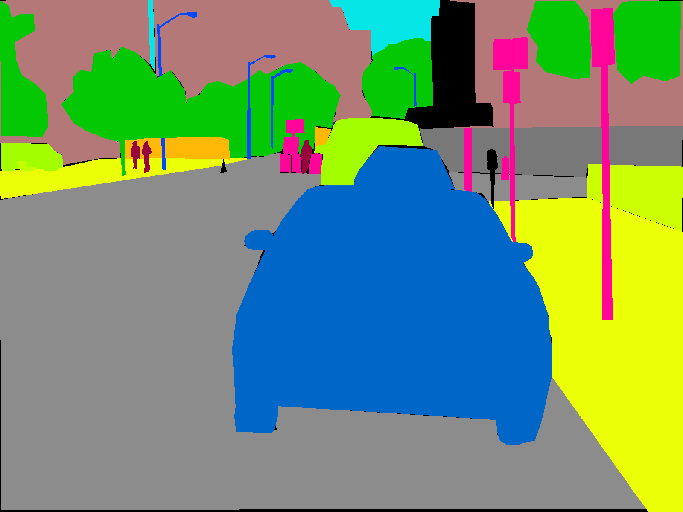} &
\includegraphics[width=\mysizeWith\linewidth]{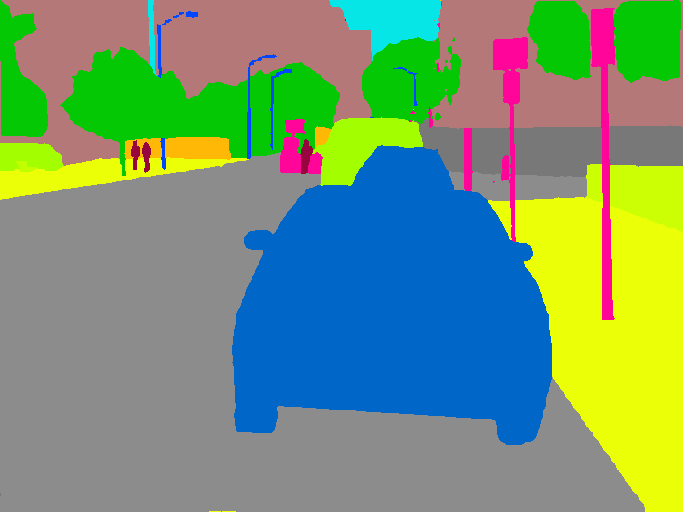} &
\includegraphics[width=\mysizeWith\linewidth]{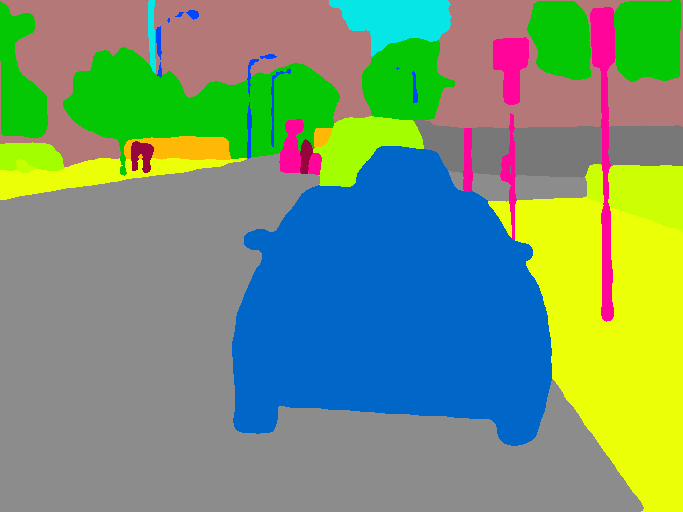} &
\includegraphics[width=\mysizeWith\linewidth]{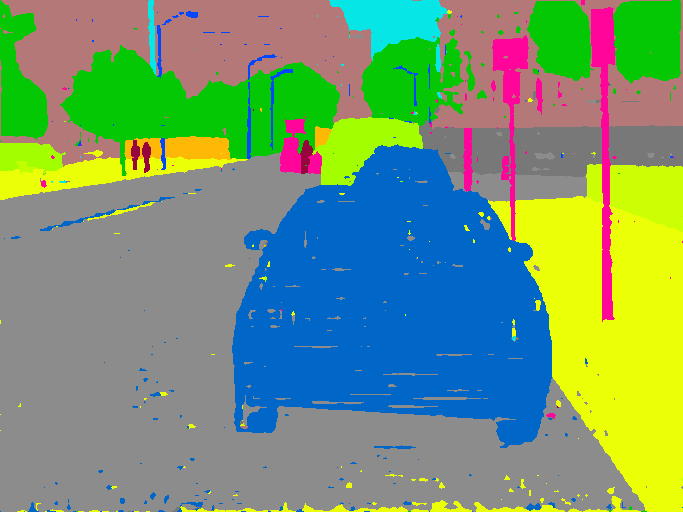} &
\includegraphics[width=\mysizeWith\linewidth]{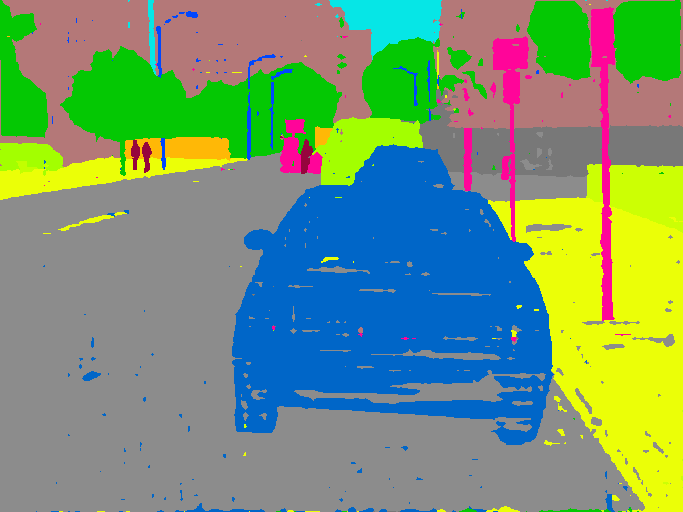} \\

\includegraphics[width=\mysizeWith\linewidth]{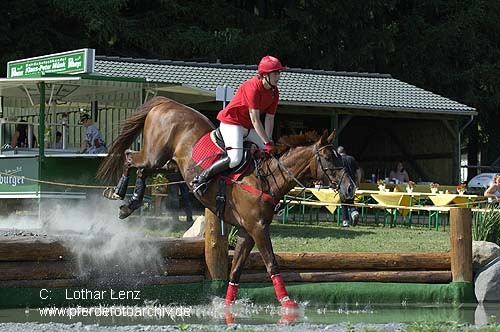} &
\includegraphics[width=\mysizeWith\linewidth]{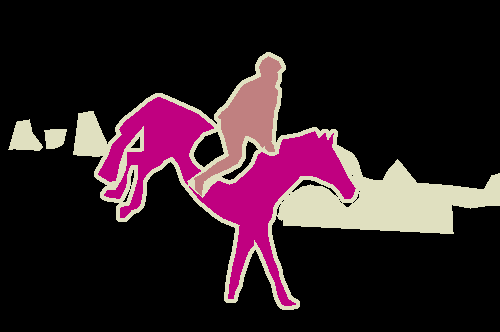} &
\includegraphics[width=\mysizeWith\linewidth]{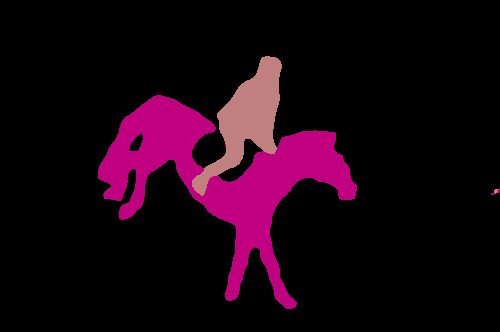} &
\includegraphics[width=\mysizeWith\linewidth]{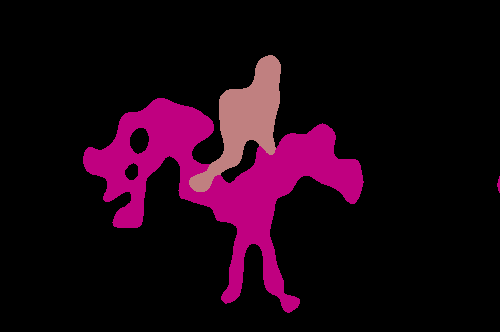} &
\includegraphics[width=\mysizeWith\linewidth]{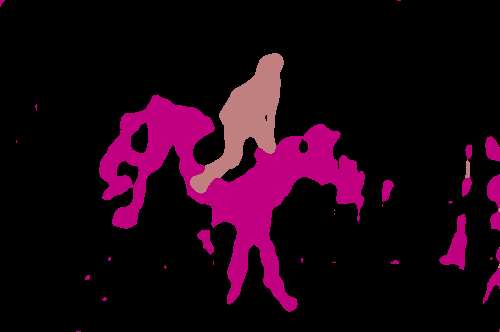} &
\includegraphics[width=\mysizeWith\linewidth]{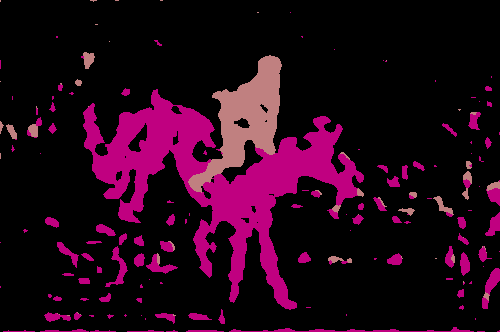} \\

\includegraphics[width=\mysizeWith\linewidth]{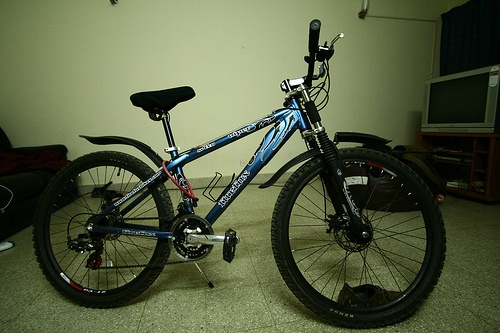} &
\includegraphics[width=\mysizeWith\linewidth]{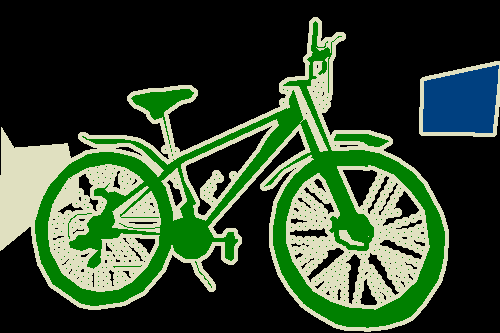} &
\includegraphics[width=\mysizeWith\linewidth]{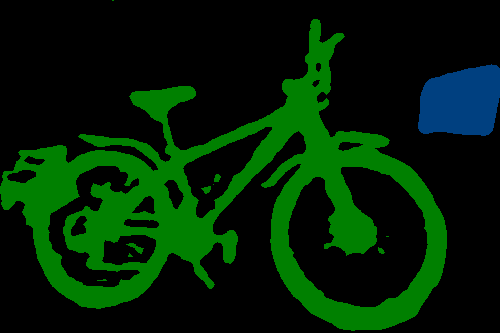} &
\includegraphics[width=\mysizeWith\linewidth]{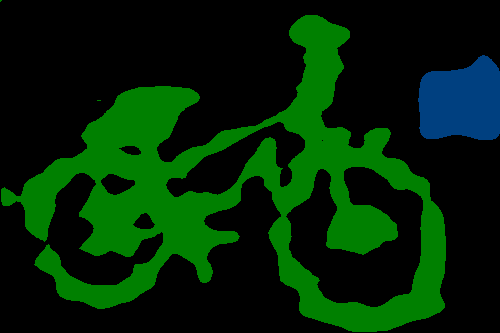} &
\includegraphics[width=\mysizeWith\linewidth]{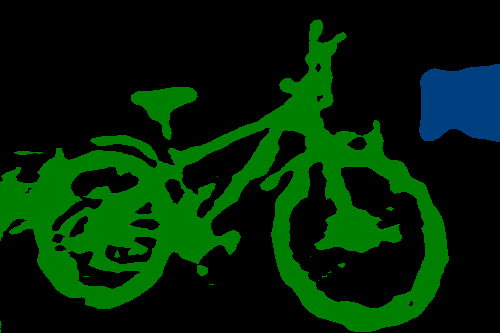} &
\includegraphics[width=\mysizeWith\linewidth]{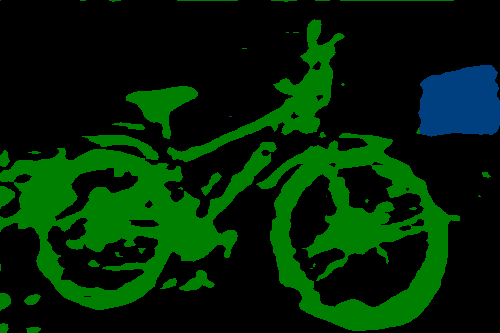} \\

\end{tabular}
\caption{{\bf Single-Image Order Training (pre-trained):} Results for different training techniques using networks with \textbf{pre-trained} (ImageNet) weights initialization. Each image represents a different network prediction, using four different network training procedures over a sparse 0.1\% labeled image.}
\label{fig:results::different_order_inference_with_pretrain}
\end{center}
\end{figure*}

\newcommand{\mysizePtsVisWithoutPreTrain}{0.23}

\begin{figure*}[htb!]
\begin{center}
\begin{tabular}{c c c}

\multirow{2}{*}{\textbf{Sampled Points}}  & \multirow{2}{*}{\textbf{Sampled Points}} & \textbf{Resulting}\\
\textbf{on Image}      &     \textbf{on Labels}           &  \textbf{Prediction}\\
\includegraphics[width=\mysizePtsVisWithoutPreTrain\linewidth]{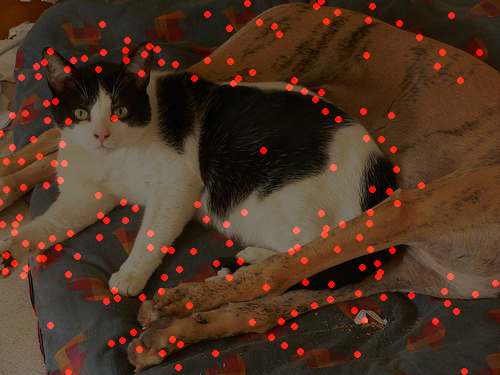} & 

\includegraphics[width=\mysizePtsVisWithoutPreTrain\linewidth]{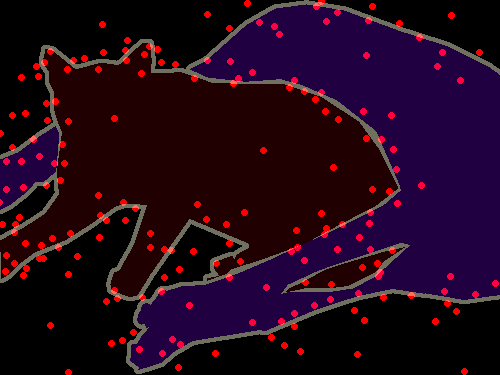} & 
\includegraphics[width=\mysizePtsVisWithoutPreTrain\linewidth]{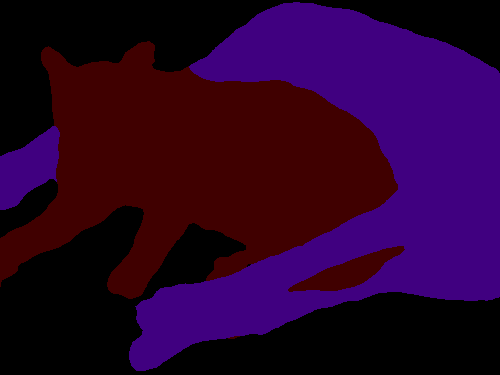} \\

\includegraphics[width=\mysizePtsVisWithoutPreTrain\linewidth]{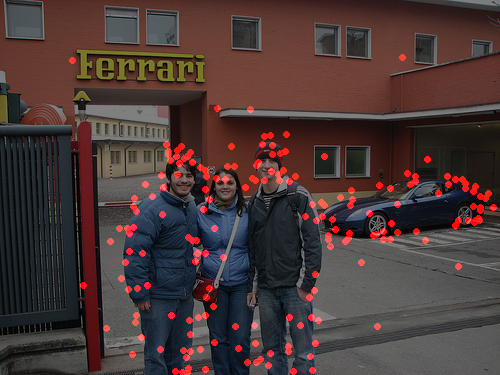} & 
\includegraphics[width=\mysizePtsVisWithoutPreTrain\linewidth]{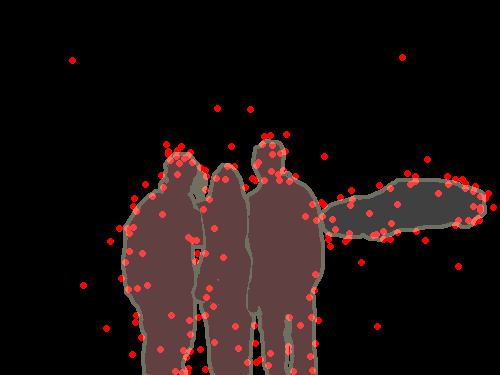} & 
\includegraphics[width=\mysizePtsVisWithoutPreTrain\linewidth]{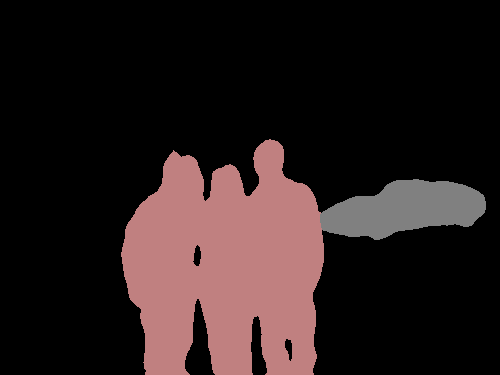} \\

\includegraphics[width=\mysizePtsVisWithoutPreTrain\linewidth]{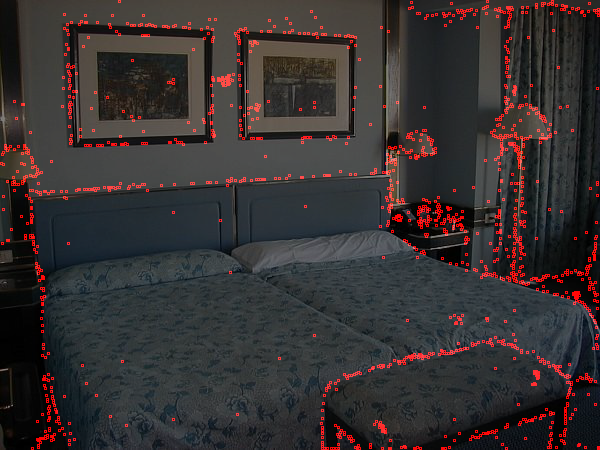} & 
\includegraphics[width=\mysizePtsVisWithoutPreTrain\linewidth]{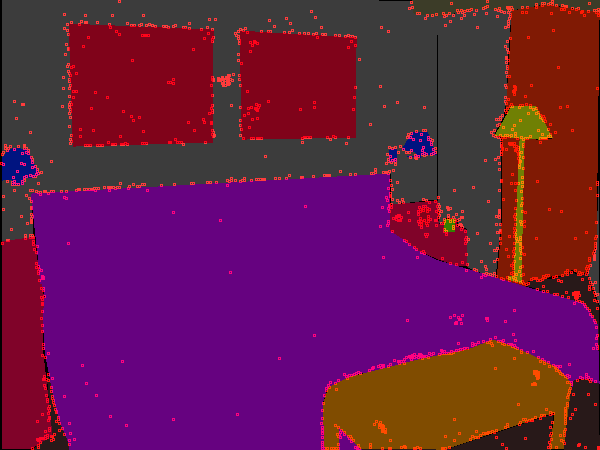} & 
\includegraphics[width=\mysizePtsVisWithoutPreTrain\linewidth]{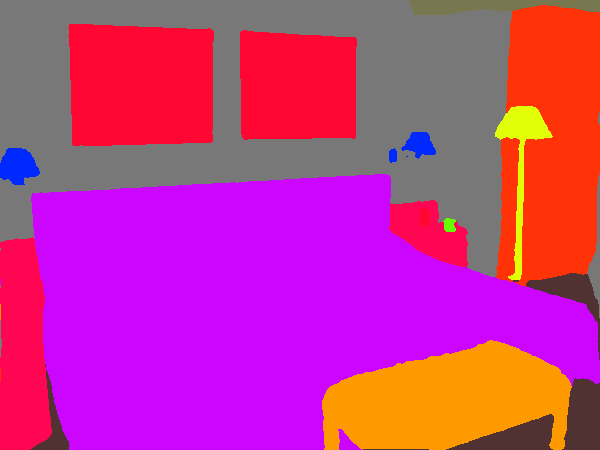} \\

\includegraphics[width=\mysizePtsVisWithoutPreTrain\linewidth]{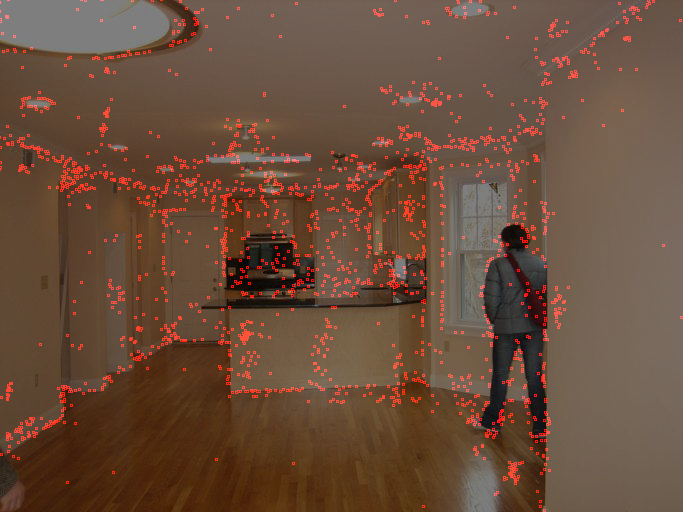} & 
\includegraphics[width=\mysizePtsVisWithoutPreTrain\linewidth]{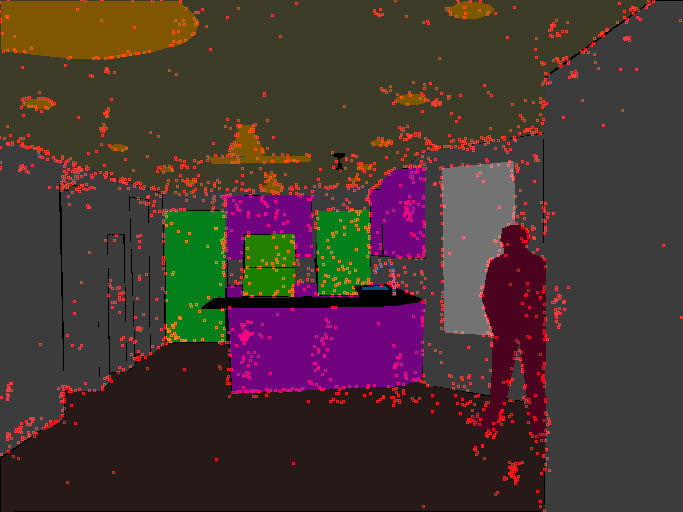} & 
\includegraphics[width=\mysizePtsVisWithoutPreTrain\linewidth]{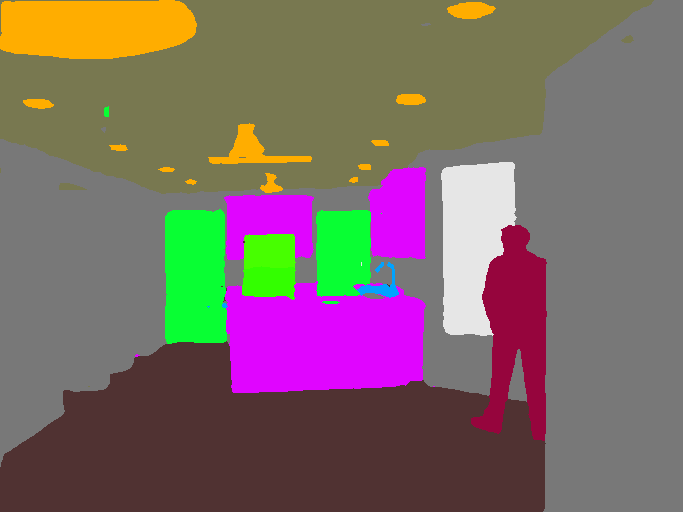} \\

\includegraphics[width=\mysizePtsVisWithoutPreTrain\linewidth]{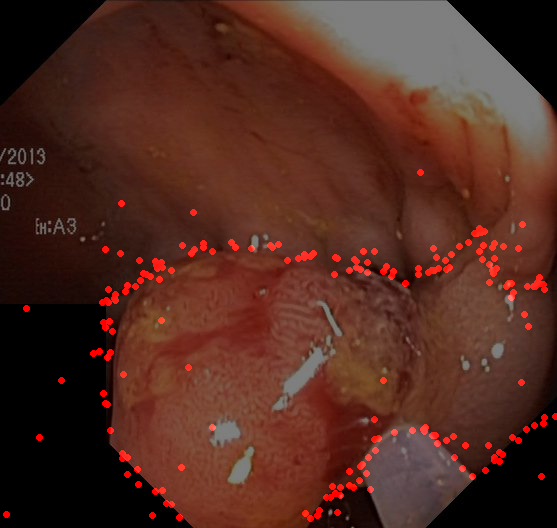} & 
\includegraphics[width=\mysizePtsVisWithoutPreTrain\linewidth]{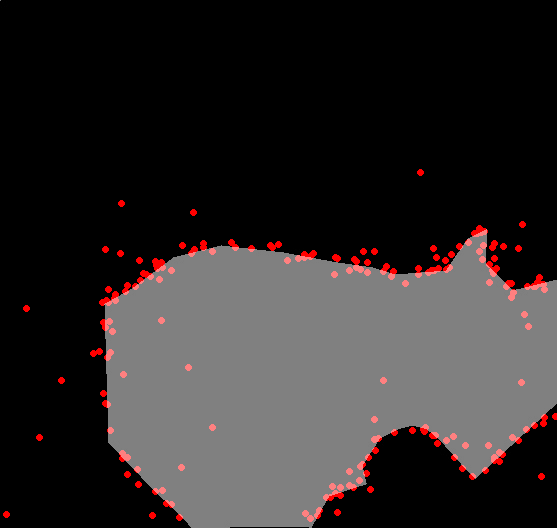} & 
\includegraphics[width=\mysizePtsVisWithoutPreTrain\linewidth]{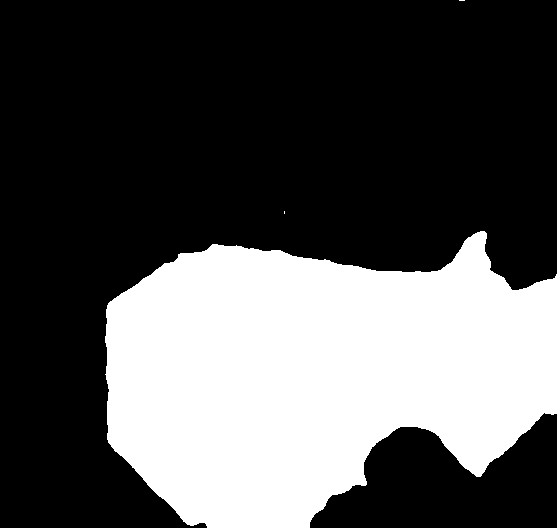} \\

\includegraphics[width=\mysizePtsVisWithoutPreTrain\linewidth]{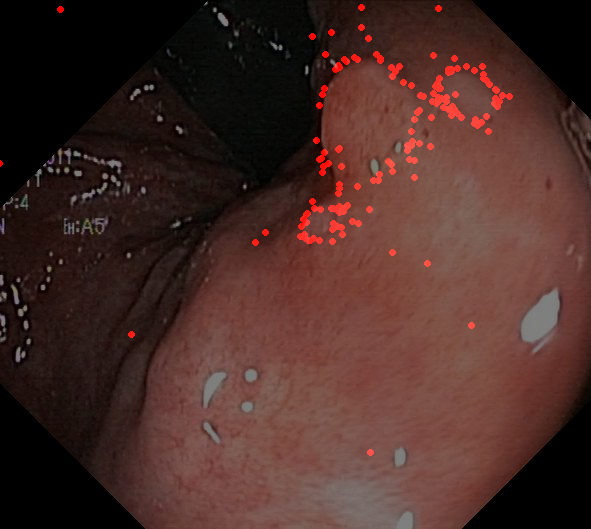} & 
\includegraphics[width=\mysizePtsVisWithoutPreTrain\linewidth]{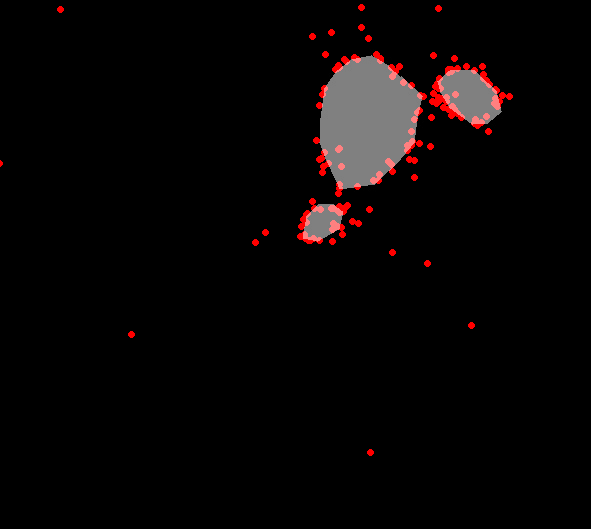} & 
\includegraphics[width=\mysizePtsVisWithoutPreTrain\linewidth]{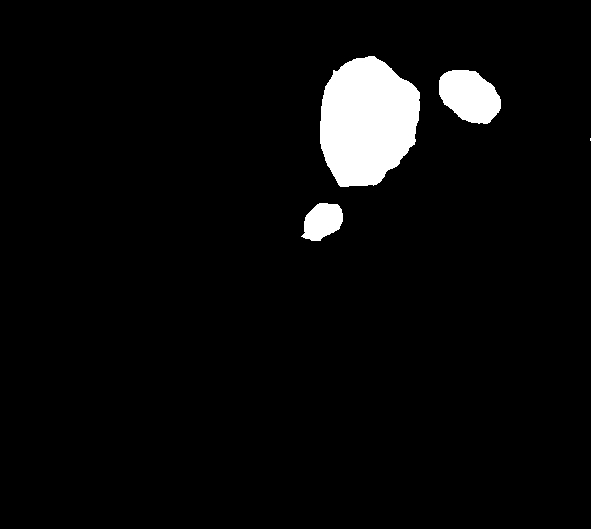} \\

\end{tabular}
\caption{{\bf Single-Image Sampled Points and Prediction (random):} Visualization of the sampled points overlaid on the image and labels and the resulting network prediction. For each image we trained a network from scratch, using only the labeled points as supervision. Each network was initialized with \textbf{random} weights.}

\label{fig:sampled_pts_visualization_without_pre_train}

\end{center}
\end{figure*}

\newcommand{\mysizePtsVisWithPreTrain}{0.21}

\begin{figure*}[htb!]
\begin{center}
\begin{tabular}{c c c}

\multirow{2}{*}{\textbf{Sampled Points}}  & \multirow{2}{*}{\textbf{Sampled Points}} & \textbf{Resulting}\\
\textbf{on Image}      &     \textbf{on Labels}           &  \textbf{Prediction}\\
\includegraphics[width=\mysizePtsVisWithPreTrain\linewidth]{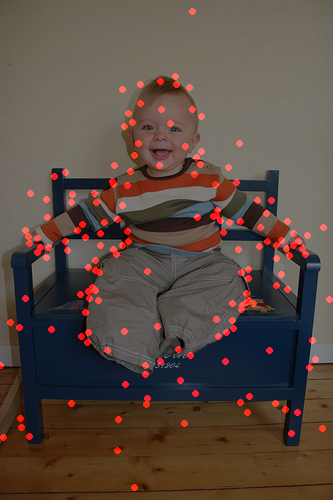} & 
\includegraphics[width=\mysizePtsVisWithPreTrain\linewidth]{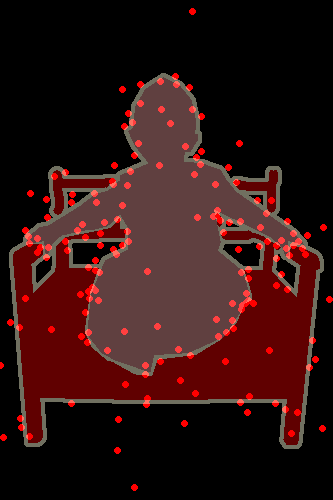} & 
\includegraphics[width=\mysizePtsVisWithPreTrain\linewidth]{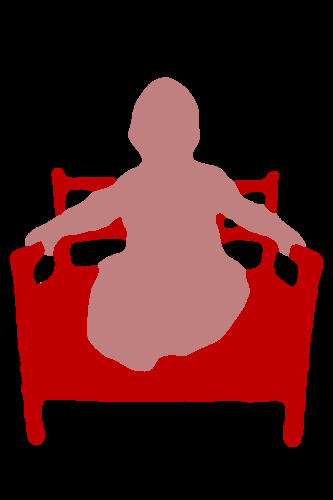} \\

\includegraphics[width=\mysizePtsVisWithPreTrain\linewidth]{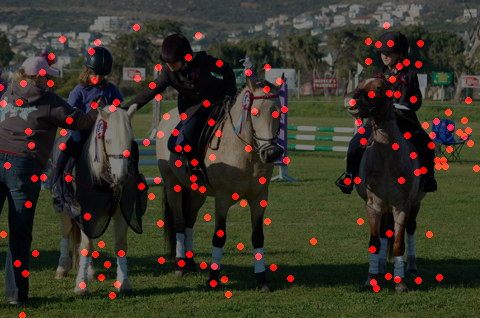} & 
\includegraphics[width=\mysizePtsVisWithPreTrain\linewidth]{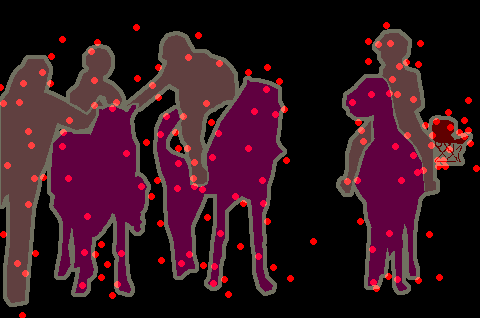} & 
\includegraphics[width=\mysizePtsVisWithPreTrain\linewidth]{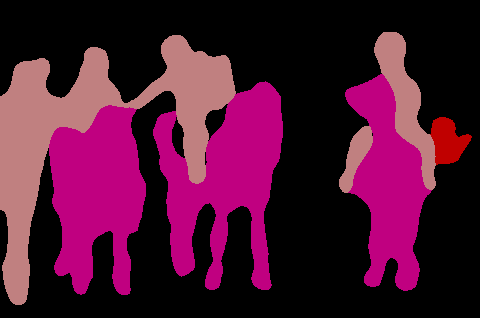} \\

\includegraphics[width=\mysizePtsVisWithPreTrain\linewidth]{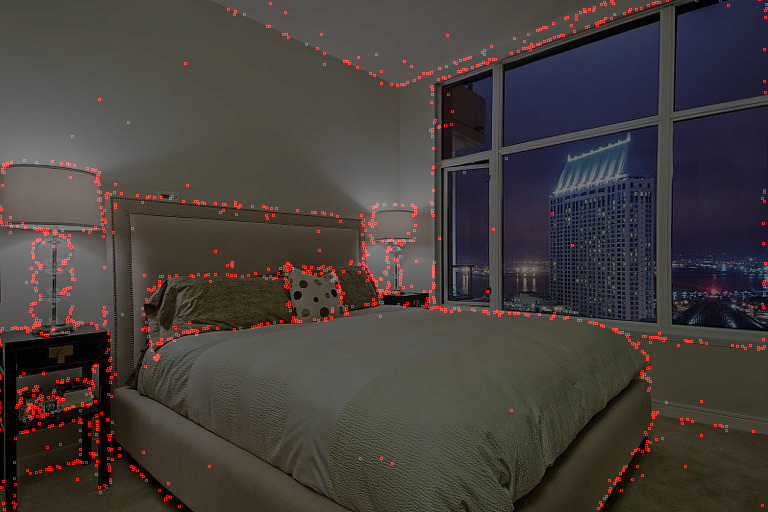} & 
\includegraphics[width=\mysizePtsVisWithPreTrain\linewidth]{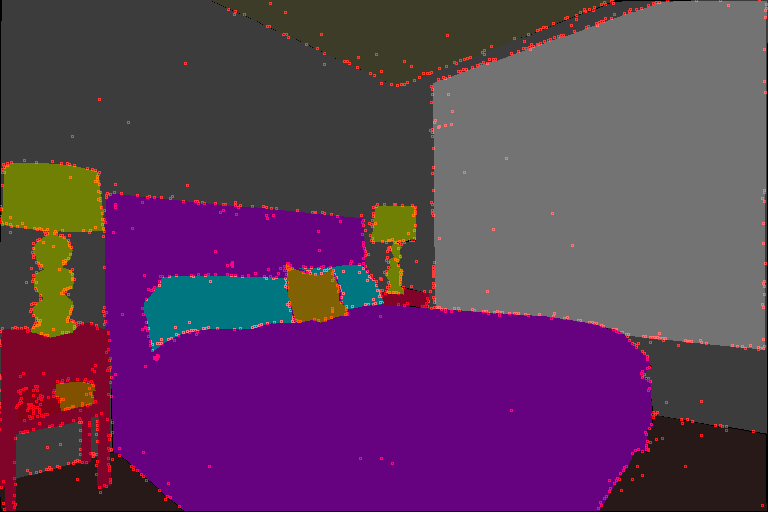} & 
\includegraphics[width=\mysizePtsVisWithPreTrain\linewidth]{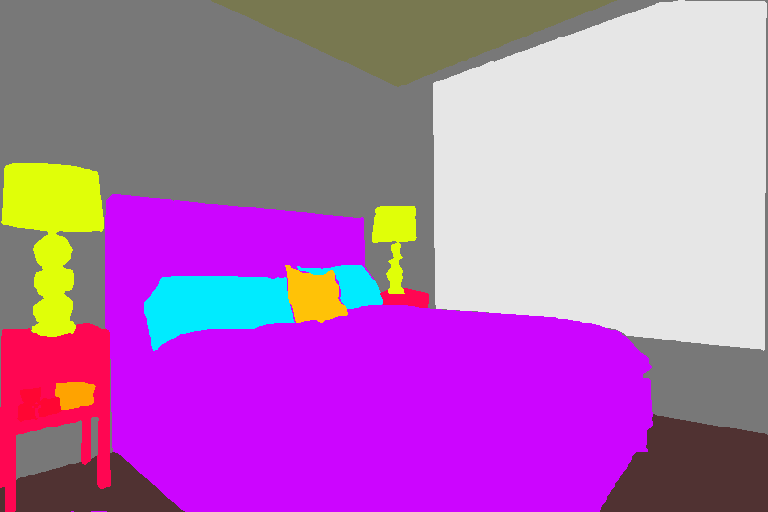} \\

\includegraphics[width=\mysizePtsVisWithPreTrain\linewidth]{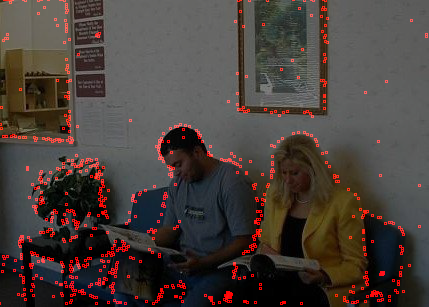} & 
\includegraphics[width=\mysizePtsVisWithPreTrain\linewidth]{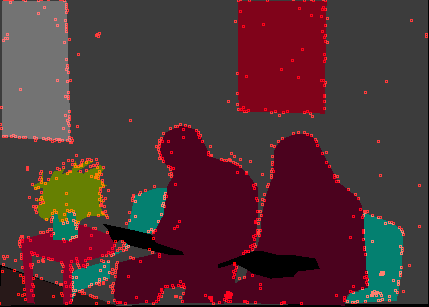} & 
\includegraphics[width=\mysizePtsVisWithPreTrain\linewidth]{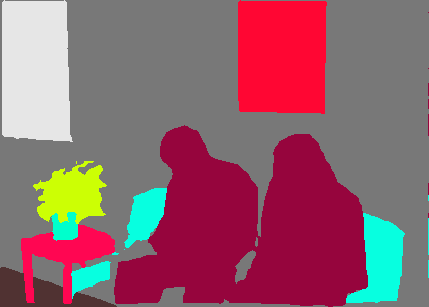} \\

\includegraphics[width=\mysizePtsVisWithPreTrain\linewidth]{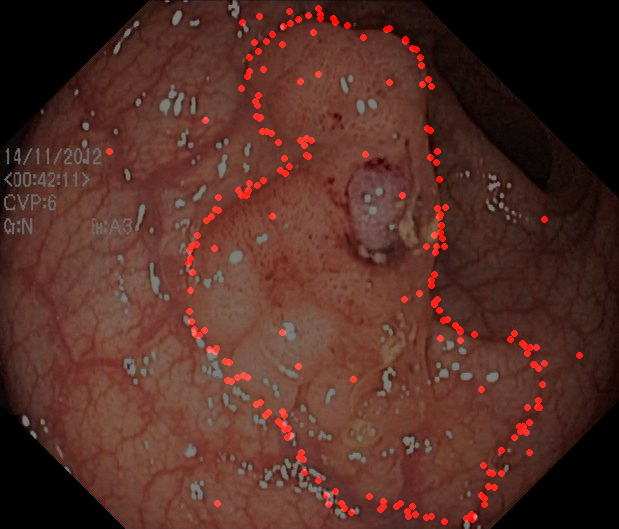} & 
\includegraphics[width=\mysizePtsVisWithPreTrain\linewidth]{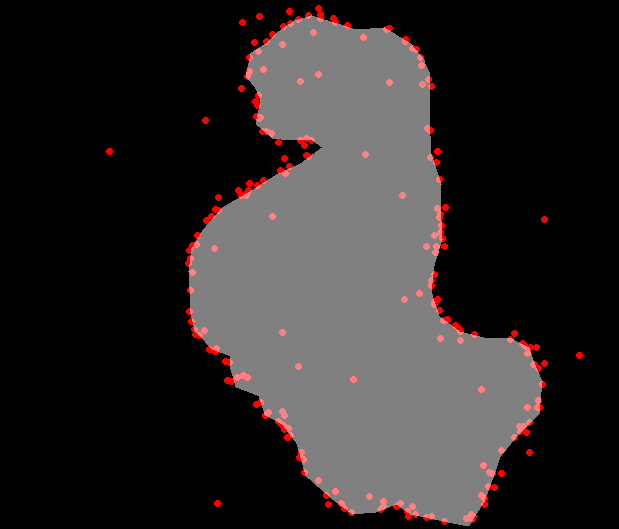} & 
\includegraphics[width=\mysizePtsVisWithPreTrain\linewidth]{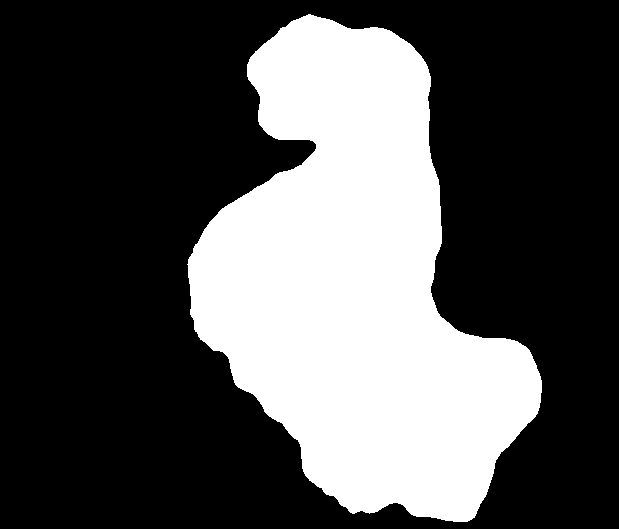} \\

\includegraphics[width=\mysizePtsVisWithPreTrain\linewidth]{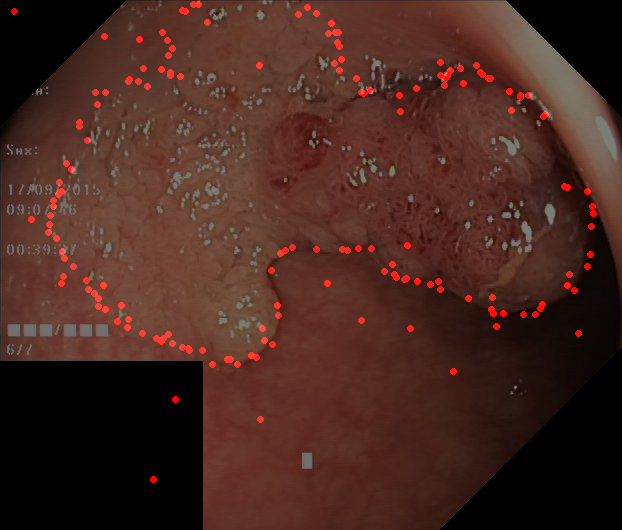} & 
\includegraphics[width=\mysizePtsVisWithPreTrain\linewidth]{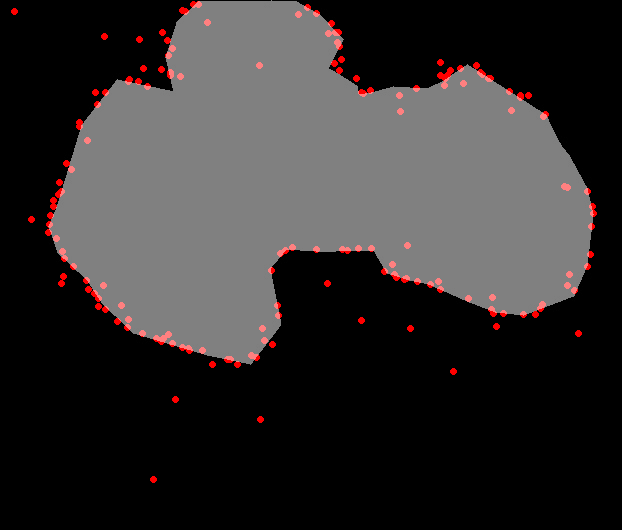} & 
\includegraphics[width=\mysizePtsVisWithPreTrain\linewidth]{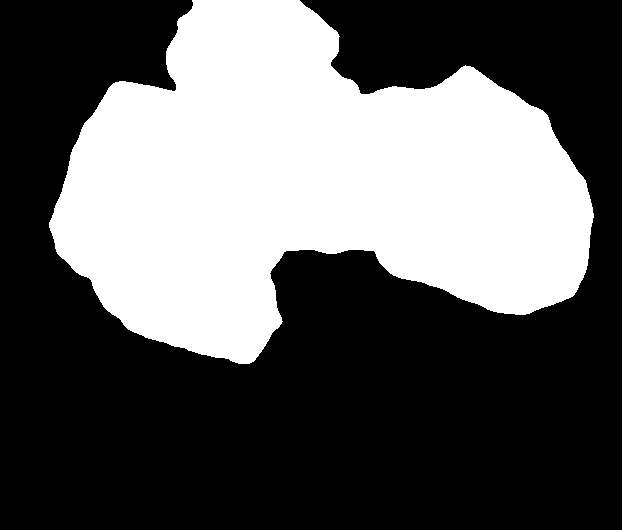} \\

\end{tabular}
\caption{{\bf Single-Image Sampled Points and Prediction (pre-trained):} Visualization of the sampled points overlaid on the image and labels and the resulting network prediction. For each image we trained a network from scratch, using only the labeled points as supervision. Each network was initialized with \textbf{pre-trained} (ImageNet) weights.}

\label{fig:sampled_pts_visualization_with_pre_train}

\end{center}
\end{figure*}


\end{document}